\documentclass[10pt,twocolumn,letterpaper]{article}

\usepackage{cvpr}              

\usepackage{graphicx}
\usepackage{amsmath}
\usepackage{amssymb}
\usepackage[pagebackref,breaklinks,colorlinks]{hyperref}
\usepackage[capitalize]{cleveref}
\crefname{section}{Sec.}{Secs.}
\Crefname{section}{Section}{Sections}
\Crefname{table}{Table}{Tables}
\crefname{table}{Tab.}{Tabs.}

\usepackage[utf8]{inputenc} 
\usepackage[T1]{fontenc}    
\usepackage{url}            
\usepackage{amsfonts}       
\usepackage{nicefrac}       
\usepackage{microtype}      
\usepackage{multirow}

\usepackage[table]{xcolor}
\usepackage[numbers]{natbib}
\usepackage{bbm}
\usepackage{amsthm}
\newtheorem{theorem}{Theorem}
\newtheorem*{theorem*}{Theorem}

\usepackage[font=small]{caption}
\usepackage{arydshln}
\theoremstyle{definition}
\newtheorem{definition}{Definition}[section]
\usepackage[linesnumbered,ruled,vlined]{algorithm2e} %
\usepackage{soul}
\usepackage{import}
\usepackage{booktabs}
\usepackage{pifont}
\usepackage{tabularx}
\usepackage{enumitem}

\usepackage[normalem]{ulem}
\newif\ifmodify
\definecolor{LightCyan}{rgb}{0.88,1,1}

\ifmodify
\newcommand{\cross}[1]{\textcolor{red}{\sout{#1}}}
\newcommand{\dl}[1]{\textcolor{red}{\sout{#1}}}

\newcommand{\f}[1]{\textcolor{orange}{#1}}

\newcommand{\zj}[1]{\textcolor{blue}{#1}}
\newcommand{\yi}[1]{\textcolor{magenta}{#1}}
\else
\newcommand{\cross}[1]{}
\newcommand{\dl}[1]{}

\newcommand{\f}[1]{#1}

\newcommand{\zj}[1]{#1}
\newcommand{\yi}[1]{#1}
\fi

\def\cvprPaperID{4954} 
\def\confName{CVPR}
\def\confYear{2022}

\begin{document}

\title{CHEX: CHannel EXploration for CNN Model Compression}


\author{
Zejiang Hou$^1$\thanks{Work done during an internship at DAMO Academy, Alibaba Group.}
\and
Minghai Qin$^2$
\and
Fei Sun$^2$
\and
Xiaolong Ma$^3$\thanks{Work partially done during an internship at Alibaba Group.}
\and
Kun Yuan$^2$
\and
Yi Xu$^4$\thanks{Work partially done during working period at Alibaba Group.}
\and
Yen-Kuang Chen$^2$
\and
Rong Jin$^2$
\and
Yuan Xie$^2$
\and
Sun-Yuan Kung$^1$
\and
\centerline{$^1$Princeton University~$^2$Alibaba Group~$^3$Northeastern University~$^4$Dalian University of Technology}
}
\maketitle

\begin{abstract}

Channel pruning has been broadly recognized as an effective technique to reduce the computation and memory cost of deep convolutional neural networks.
However, conventional pruning methods have limitations in that: they are restricted to pruning process only, and they require a fully pre-trained large model. Such limitations may lead to sub-optimal model quality as well as excessive memory and training cost.
In this paper, we propose a novel Channel Exploration methodology, dubbed as CHEX, to rectify these problems.
As opposed to pruning-only strategy, we propose to repeatedly prune and regrow the channels throughout the training process, which reduces the risk of pruning important channels prematurely. More exactly:
From intra-layer’s aspect, we tackle the channel pruning problem via a well-known column subset selection (CSS) formulation.
From inter-layer’s aspect, our regrowing stages open a path for dynamically re-allocating the number of channels across all the layers under a global channel sparsity constraint 
.
In addition, all the exploration process is done in a single training from scratch without the need of a pre-trained large model.
Experimental results demonstrate that CHEX can effectively reduce the FLOPs of diverse CNN architectures on a variety of computer vision tasks, including image classification, object detection, instance segmentation, and 3D vision. 
For example, our compressed ResNet-50 model on ImageNet dataset achieves 76\% top-1 accuracy with only 25\% FLOPs of the original ResNet-50 model, outperforming previous state-of-the-art channel pruning methods. {
The checkpoints and code are available at \href{https://github.com/zejiangh/Filter-GaP}{\text{here}}
}.

\end{abstract}

\section{Introduction}
Albeit the empirical success of deep convolutional neural networks (CNN) on many computer vision tasks, the excessive computational and memory cost impede their deployment on {mobile or edge devices}.
Therefore, it is vital to explore model compression, which reduces the redundancy {in the} {model} while {maximally} maintaining the accuracy.

Among various model compression approaches, channel pruning
has been recognized as an effective tool to achieve practical memory saving and inference acceleration on general-purpose hardware.
To derive a sub-model, channel pruning removes the redundant channels along with all the associated filters connected to those channels. 

\begin{figure}[t]
\centering
\includegraphics[scale=0.29]{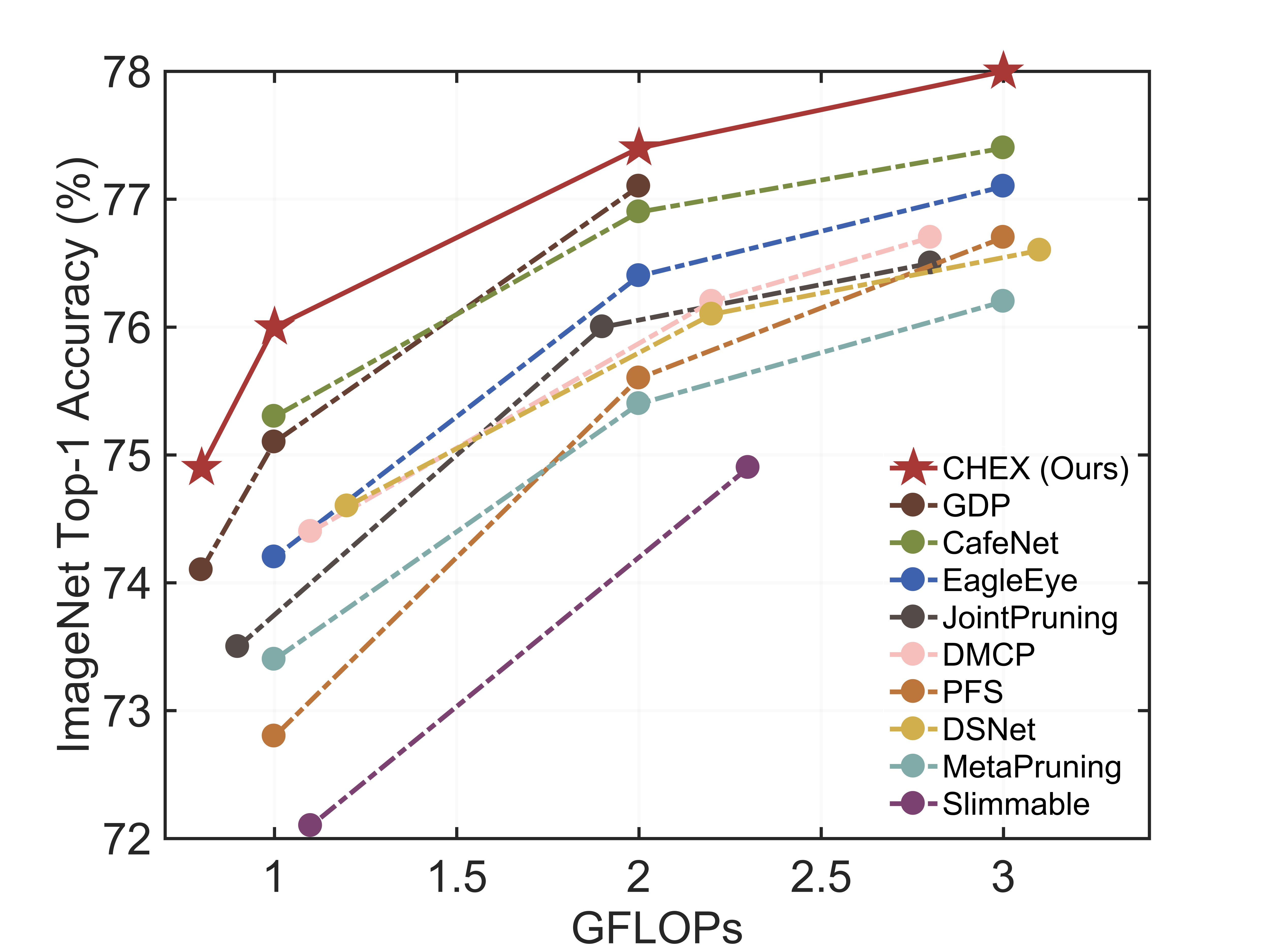}
\caption{Comparison of {the} accuracy-FLOPs Pareto curve of {the} compressed ResNet-50 {models} on ImageNet. CHEX shows the top-performing Pareto frontier compared with previous {methods.} And we obtain the sub-models without pre-training a large model.}
\label{fig:pareto}
\vspace{-0.1in}
\end{figure}


\begin{figure*}[t]
\centering
\includegraphics[scale=0.16]{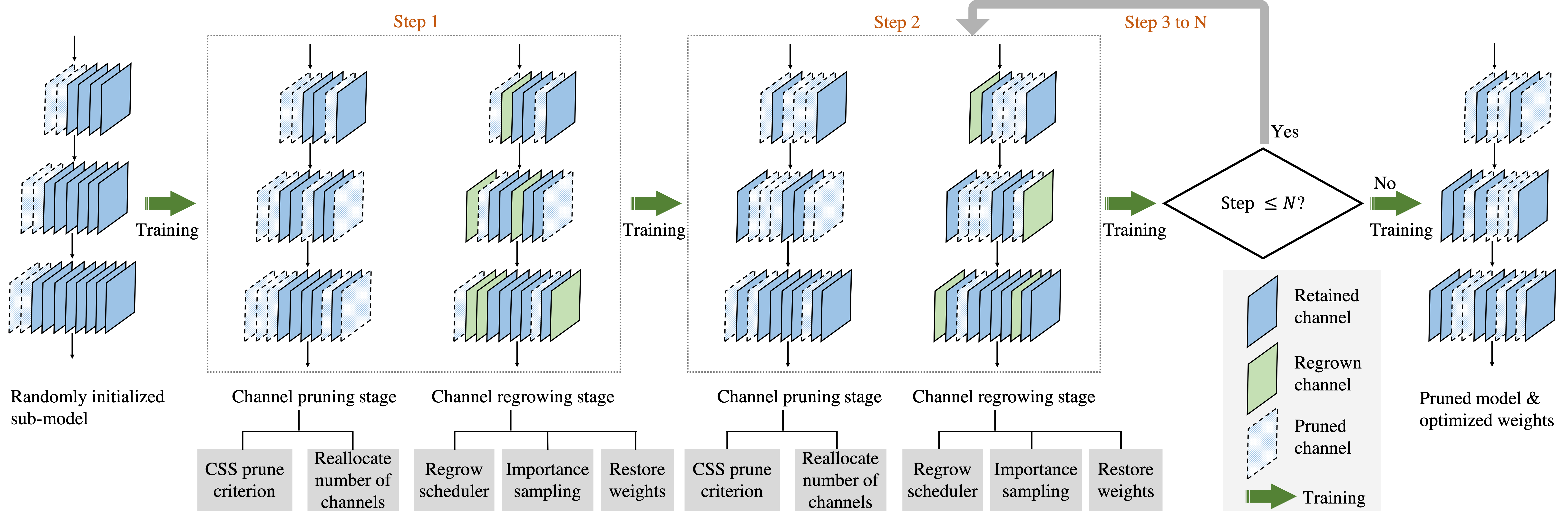}
\caption{An illustration of our CHEX method, which jointly optimizes the weight values and explores the sub-model structure in one training pass from scratch. In CHEX, both retained and regrown channels in the sub-model are active, participating in the training iterations.}
\label{figure 1}
\vspace{-0.1in}
\end{figure*}

Most existing channel pruning methods~\cite{li2016pruning,he2017channel, luo2017thinet, yang2018netadapt, zhuang2018discrimination, chin2019legr, he2019filter, you2019gate, molchanov2019importance, guo2020channel,  ye2020good, hou2020feature, hou2020efficient, ma2021non, zhang2021structadmm}
adopt a progressive pruning-training pipeline: pre-training a large model until convergence, pruning a few unimportant channels by the pre-defined criterion, and finetuning the pruned model to restore accuracy. 
The last two stages are usually executed in an interleaved manner repeatedly, 
which suffers from long training time.
Various attempts have been made to improve the pruning efficiency. 
Training-based channel pruning methods~\cite{liu2017learning,lin2019towards,li2019compressing,guo2020multi,li2020group,zhuang2020neuron} impose sparse regularization {such as LASSO or group LASSO} to the model parameters during training.
However, these commonly adopted regularization may not penalize the parameters to exactly zero. Removing many small but non-zero parameters inevitably damages the model accuracy. 
Although this problem can be addressed by applying specialized optimizers~\cite{huang2018data,ding2019centripetal} or model reparameterization tricks~\cite{ding2020lossless}, these methods require a well-pretrained large model, which counters our goal of improving pruning efficiency.
Sampling-based methods~\cite{YuanSM21, ning2020dsa, kang2020operation, gao2020discrete, louizos2017learning, hou2022multi} directly train sparse models. 
However, these methods may suffer from training instability and converge to sub-optimal points~\cite{guo2021gdp}.
\cite{wang2019pruning,you2019drawing,lee2018snip,wang2020picking} shorten or eliminate the pre-training phase, and extract the sub-model at an early stage or even from random initialization. These methods are incapable to recover prematurely pruned important channels, which limits the model capacity and leads to unacceptable accuracy degradation.

To rectify the aforementioned limitations, we propose a \textit{channel exploration} methodology called \textbf{CHEX} to obtain high accuracy sub-models without pre-training a large model or finetuning the pruned model.
In contrast to {the} traditional pruning {approaches} that permanently remove channels~\cite{li2016pruning, luo2017thinet, yu2018nisp, zhuang2018discrimination},
we dynamically adjust the importance of \f{the} channels
via {a} \textit{periodic pruning and regrowing process}, which allows 
{the prematurely pruned channels to be recovered} and prevents the model from losing the representation ability early in the training process. 
From intra-layer's perspective, we re-formulate the channel pruning problem into the classic \textit{column subset selection} (CSS) problem in linear algebra, leading to a closed-form solution.
From inter-layer's perspective, rather than sticking to a fixed or manually designed sub-model structure, our approach re-evaluates the importance of different layers after each regrowing stage. This leads to a \textit{sub-model structure exploration} technique to dynamically re-allocate the number of channels across all the layers under a given budget.
With only one training pass from scratch, our obtained sub-models yield better accuracy than the previous methods under the same FLOPs reductions. 
Moreover, the simplicity of our method allows us to derive a theoretical analysis on the training behaviour of CHEX, to provide further insights and interpretation.


The contributions of this paper are highlighted as follows: 
\begin{itemize}[noitemsep]
    \item We propose a channel exploration method CHEX 
    with three novel features: (1) a periodic channel pruning and regrowing process, (2) a pruning criterion (i.e., leverage score) based on column subset selection, and (3) a sub-model structure exploration technique.
    \item Our method obtains the sub-model in one training pass from scratch, effectively reducing the training cost, because it circumvents the expensive pretrain-prune-finetune cycles.
    \item Experimentally, CHEX exhibits superior accuracy under different FLOPs constraints (as shown in Figure~\ref{fig:pareto}), and is applicable to a plethora of computer vision tasks. For image classification on ImageNet, our compressed ResNet-50 model yields 4$\times$ FLOPs reduction while achieving 76\% top-1 accuracy, improving previous best result~\cite{su2021locally} by 0.7\%.
    For object detection on COCO dataset, our method achieves 2$\times$ FLOPs reduction on the SSD model while improving $0.7\%$ mAP over the unpruned baseline. 
    For instance segmentation on COCO dataset and 3D point cloud segmentation on ShapeNet dataset, our method achieves 2$\times$ and 11$\times$ FLOPs reductions {on the} Mask R-CNN and PointNet++ {models} with negligible {quality} loss compared to the unpruned models, respectively.
    \item {We provide a theoretical convergence guarantee of our CHEX method from the view of non-convex optimization setting, which is applicable to deep learning.}
\end{itemize}

\section{Methodology}


Our method takes an existing CNN model as our channel exploration space, which provides the maximum number of explored channels.
As illustrated in Figure~\ref{figure 1}, we describe the training flow of CHEX using a 3-layer CNN model as an example, in which we set the target channel sparsity\footnote{We {follow} the notion of structured sparsity {introduced in}~\cite{wen2016learning}. Our sub-model is a slimmer dense CNN model.} to 50\%. From top to bottom, the three layers contain 6, 8, and 10 channels, respectively, denoted as $\text{Conv}-6-8-10$.
Our method starts with a randomly initialized sub-model.
During training, at periodically spaced training iterations (e.g., pre-determined $\Delta T$ iterations), a channel pruning and regrowing process is performed, which is referred to as one step in CHEX. A total of $N$ steps are applied, and each step is composed of the following stages:
\begin{itemize}
    \item A pruning stage removes the unimportant channels to the target sparsity.
    The number of channels are re-allocated across all different layers via the sub-model structure exploration technique.
    For example, in Figure~\ref{figure 1}, the pruned sub-model in step 1 has an architecture $\text{Conv}-3-4-5$, retaining 12 out of 24 channels. It may be adjusted later to $\text{Conv}-2-3-7$ in step 2. Such adaptations continue in the loop across the $N$ steps.
    
    \item Immediately after pruning, a channel regrowing stage regrows back a fraction of {the} previously pruned channels, whose weight values are restored to their most recently used values {before being pruned}. Note that the 
    {regrown} channels {may be} pruned multiple steps before.
    A decay scheduler is adopted to gradually reduce the number of regrown channels across the $N$ steps. {For example, in Figure~\ref{figure 1}, the channel regrowing stage in step 1 re-activates six channels, while it only re-activates four channels in step 2.} 
\end{itemize}

Our method interleaves the model training and the periodic pruning-regrowing stages. Since the total number of regrown channels is reduced at each step, the sub-model under training enjoys gradually decreased computation cost and converges to the target channel sparsity in the end.
As an algorithm guideline, the pseudo-code of CHEX is provided in Algorithm~\ref{algorithm: overview}.

\setlength{\textfloatsep}{0.1cm}
\begin{algorithm}[t]
\small
    \textbf{Input}: An $L$-layer CNN model with weights $\mathbf{W}=\{\mathbf{w}^1,...,\mathbf{w}^L\}$; target channel sparsity $S$; total training iterations $T_{\text{total}}$; initial regrowing factor $\delta_0$; training iterations between two consecutive steps $\Delta T$; total pruning-regrowing steps $T_\text{max}$; training set $\mathcal{D}$ \;
    \textbf{Output}: A sub-model satisfying the target sparsity $S$ and its optimal weight values $\mathbf{W}^*$\;
    Randomly initialize the model weights $\mathbf{W}$\;
    \For{each training iteration $t\in[T_\text{total}]$}
    {
    Sample a mini-batch from $\mathcal{D}$ and update the model weights $\mathbf{W}$ \;
    \If{$\text{Mod}(t,\Delta T)=0$ and $t\leq T_\text{max}$} 
    {
    Re-allocate the number of channels for each layer in the sub-model $\{\kappa^l,l\in[L]\}$ by Eq.\eqref{layer-wise sparsity} \;
    Prune $\{\kappa^lC^l,l\in[L]\}$ channels by CSS-based pruning in Algorithm~\ref{algorithm: CSS-based filter pruning} \; 
    Compute the channel regrowing factor by a decay scheduler function \;
    Perform importance sampling-based channel regrowing in Algorithm~\ref{algorithm: Sampling-based filter regrowing} \;
    }
    }
\caption{Overview of the CHEX method.}
\label{algorithm: overview}
\end{algorithm}
\setlength{\floatsep}{0.1cm}

\subsection{Channel pruning stage}
\label{pruning}
Suppose a CNN {contains} $L$ layers with parameters $\{\mathbf{w}^1,...,\mathbf{w}^L\}$, {with} $\mathbf{w}^l\in\mathbb{R}^{H^lW^lC^{l-1}\times C^l}$ denoting the reshaped\footnote{{For ease of derivation, we convert the convolution weight tensor of size $H^l\times W^l \times C^{l-1} \times C^l$ to {the} matrix of size $H^lW^lC^{l-1}\times C^{l}$, where the channels are listed as columns in the reshaped weight matrix.}} convolution weight matrix.
$C^{l-1}, C^l, H^l\times W^l$ represent {the number of} input channels, {the} number of output channels, {and the} kernel size, respectively.
{The} $j$-th channel in {the} $l$-th layer {is denoted as} $\mathbf{w}^l_{:,j}$. That is, a column in $\mathbf{w}^l$ represents one channel in the convolution.
For notation simplicity, we use $K^l=H^lW^lC^{l-1}$ in the following text, i.e., $\mathbf{w}^l\in\mathbb{R}^{{K^l}\times C^l}$.

Suppose $\kappa^l$ denotes the {channel sparsity of the} $l$-th layer.
For each layer, channel pruning identifies a set of important channels with index set $\mathcal{T}^l$ ($|\mathcal{T}^l|=\lceil(1-\kappa^l)C^l\rceil$), which retains the most important information in $\mathbf{w}^l$, such that the remaining ones $\{\mathbf{w}^l_{:,j}, j\notin\mathcal{T}^l\}$ {may be discarded with minimal impact to model accuracy}. In other words, channel pruning selects the most ``representative'' columns from $\mathbf{w}^l$ that can reconstruct the original weight matrix {with minimal error}.
From this perspective, channel pruning {can be naturally represented as} the \textit{Column Subset Selection} (CSS) {problem} in linear algebra~{\cite{gu1996efficient}}. {This provides} us a new theoretical guideline for designing channel pruning criterion in a principled way, rather than depending on heuristics. 
To rigorously characterize the most ``representative" columns of a matrix, we {formally define} CSS as follows:
\begin{definition}[Column Subset Selection]
{Given a} matrix ${\mathbf{w}}^l\in\mathbb{R}^{K\times C^l}$, {let} $c\leq C^l$ be the number of columns to select. Find $c$ columns of ${\mathbf{w}}^l$, denoted {by} ${\mathbf{w}}^l_c$, that would minimize:
\begin{equation}
    \|{\mathbf{w}}^l-{\mathbf{w}}^l_c{{\mathbf{w}}^{l}_c}^{\dagger}{\mathbf{w}}^l\|^2_F ~~~\text{or}~~~ \|{\mathbf{w}}^l-{\mathbf{w}}^l_c{{\mathbf{w}}^{l}_c}^{\dagger}{\mathbf{w}}^l\|^2_2\label{CSSP},
\end{equation}
where ${\dagger}$ stands for the Moore-Penrose pseudo-inverse, $\|\cdot\|_F$ and $\|\cdot\|_2$ represent matrix Frobenius norm and spectral norm, respectively.
\end{definition}

    

\begin{algorithm}[t]
\small
    \textbf{Input}: Model weights ${\mathbf{w}}^l$; pruning ratios $\kappa^l$ \;
    \textbf{Output}: \f{The} pruned layer $l$ \;
    {Compute the} number of retained channels $\tilde{C}^l=\lceil(1-\kappa^l)C^l\rceil$ \; 
    Compute {the} top $\tilde{C}^l$ right singular vectors $\mathbf{V}_{\tilde{C}^l}^l $ of ${\mathbf{w}}^l$ \;
    Compute {the} leverage scores for all the channels in layer $l$ $\psi_j^l=\|[\mathbf{V}_{\tilde{C}^l}^l]_{j,:}\|_2^2$ for all $j\in[C^l]$ \;
    Retain {the} {important} channels identified as
    $\mathcal{T}^l=\text{ArgTopK}(\{\psi_j^l\};\tilde{C}^l)$ \;
    Prune channels $\{\mathbf{w}^l_{:,j},j\notin\mathcal{T}^l\}$ from layer $l$ \;
\caption{CSS-based channel pruning.}
\label{algorithm: CSS-based filter pruning}
\end{algorithm}

{Since channel pruning and CSS share the same goal of best recovering the full matrix {by} a subset of {its} columns,} we can leverage the rich theoretical foundations of CSS to derive a new pruning criterion. Our channel pruning stage is conducted periodically during training, thus we employ a computationally efficient deterministic CSS algorithm, referred {to} as the \textit{Leverage Score Sampling}~\cite{papailiopoulos2014provable}. 
The core of this algorithm involves the leverage scores of matrix ${{\mathbf{w}}^l}$, which are defined as follows:
\begin{definition}[Leverage Scores]\label{def:LS}
Let $\mathbf{V}_c\in\mathbb{R}^{C^l\times c}$ be the top-$c$ right singular vectors of ${{\mathbf{w}}^l}$ ($c$ represents the number of {selected} columns from $\mathbf{w}^l$). Then, the leverage score of the $j$-th column of ${{\mathbf{w}}^l}$ is given as: $\psi_{j}=\|[\mathbf{V}_c]_{j,:}\|_2^2$, where $[\mathbf{V}_c]_{j,:}$ denotes the $j^{th}$ row of $\mathbf{V}_c$.
\end{definition}

The leverage score sampling algorithm samples $c$ columns of ${{\mathbf{w}}^l}$ that corresponds to the largest $c$ leverage scores of ${{\mathbf{w}}^l}$. 
Despite its simplicity, theoretical analysis in~\cite{papailiopoulos2014provable} has shown that this deterministic solution provably obtains near optimal low-rank approximation error for Eq.\eqref{CSSP}.

Based on the above analysis, we propose a CSS-based channel pruning method with leverage score sampling, as shown in Algorithm \ref{algorithm: CSS-based filter pruning}. {When} given a pruning ratio $\kappa^l$ {for layer $l$}, {we need to select and retain $\lceil(1-\kappa^l)C^l\rceil$ important channels}. We first compute the top $\lceil(1-\kappa^l)C^l\rceil$ right singular vectors of {the} weight matrix $\mathbf{w}^l$. {Then, we calculate} the leverage scores of all the channels in this layer as Definition {\ref{def:LS}}, and {rank them} in descending order.
Finally, we identify the set of important channels to retain as $\mathcal{T}^l=\text{ArgTopK}(\{\psi_j^l\};\lceil(1-\kappa^l)C^l\rceil)$,
which gives the indices of channels with the top $\lceil(1-\kappa^l)C^l\rceil$ leverage scores of $\mathbf{w}^l$. The {remaining} bottom-ranking channels {are} pruned.

\begin{algorithm}[t]
\small
    \textbf{Input}: Indices of active channels $\mathcal{T}^l$ in the sub-model; 
    regrowing factor $\delta_t$\;
    \textbf{Output}: \f{The} regrown layer $l$ \;
    {
    Compute \f{the} importance sampling probabilities by Eq.\eqref{orthogonal_projection} $p^l_j=\text{exp}(\epsilon^l_j)/\sum_{j'}\text{exp}(\epsilon^l_{j'})$ for all $j\hspace{-0.02in}\in\hspace{-0.02in}[C^l]\hspace{-0.02in}\setminus\hspace{-0.02in}\mathcal{T}^l$ \;
    Compute the number of regrown channels $k^l=\lceil \delta^tC^l \rceil$ \;
    Perform importance sampling $\mathcal{G}^l\hspace{-0.04in}=\hspace{-0.04in}\text{Multinomial}(\{p^l_j\};k^l)$ \; 
    Restore \f{the} MRU weights of the chosen channels $\{\hat{\mathbf{w}}^l_j,j\in \mathcal{G}^l\}$ \;
    }
    Regrow channels $\{\hat{\mathbf{w}}^l_j, j\in \mathcal{G}^l\}$ to layer $l$ \;
\caption{Sampling-based channel regrowing.}
\label{algorithm: Sampling-based filter regrowing}
\end{algorithm}

\subsection{Channel regrowing stage}
\label{regrow}

Since the method trains from a randomly initialized model and the pruning stage may be based on the weights that are not sufficiently trained. In the early stage of training, the pruning decisions may not be optimal and some important channels are prematurely pruned. Therefore, after {each} pruning {stage}, our method regrows \textit{a fraction} of the previously pruned channels back to the model. 
The regrown channels are updated in the subsequent training. If they are important to the model, they {may} survive the future pruning stages after a number of iterations of {training}. 
Moreover, {the} channel regrowing stage enables the model to have better representation ability during training, since the model capacity is not permanently restricted as the one-shot pruning methods.

{To complete the regrowing stage, we need to assign proper weight values to the newly activated channels. One straightforward choice is to assign zero values for stable training, since the regrown channels do not affect the output of the model. However, we find that regrowing channels with zeros would receive zero gradients in the subsequent training iterations. This is undesirable because the regrown channels would remain deactivated and the method degenerates to the one-shot early-bird pruning \cite{you2019drawing}. Based on our ablations, we find the best scheme is that the newly activated channels restore their \textit{most recently used} (MRU) parameters, which are the last values before they are pruned.} We constantly maintain a copy of the weight values of the pruned channels,
in case that they may get regrown back in the future regrowing stages. Note that the regrowing stage may regrow channels that {are} pruned multiple steps before, instead of just re-activating what are pruned at the pruning stage immediately before the current regrowing stage. 


To determine the channels to regrow, a naive way is to perform uniform sampling from the candidate set $\{j|j\in [C^l]\setminus\mathcal{T}^l\}$. 
However, uniform sampling does not consider the possible inter-channel dependency between the active channels survived in the sub-model and the candidate channels to regrow. Instead, we propose an \textit{importance sampling} strategy based on channel orthogonality for regrowing, as shown in Algorithm \ref{algorithm: Sampling-based filter regrowing}. Channel orthogonality automatically implies linear independency, which helps avoid trivial regrowing where the newly regrown channels lie in the span of the active channels. Channel orthogonality also encourages the channel diversity and improves model accuracy \cite{bansal2018can}. 
Formally, we denote the active channels in layer $l$ by matrix $\mathbf{w}^l_{\mathcal{T}^l}$. The orthogonality $\epsilon_j^l$ of a channel $\mathbf{w}_j^l$ in the candidate set with respect to the active channels can be computed by the classic orthogonal projection formula \cite{horn2012matrix}:
\begin{equation}
    \epsilon_j^l = \|\mathbf{w}_j^l - \mathbf{w}^l_{\mathcal{T}^l}(\mathbf{w}^{l^T}_{\mathcal{T}^l}\mathbf{w}^l_{\mathcal{T}^l})^{\dagger}\mathbf{w}^{l^T}_{\mathcal{T}^{l}}\mathbf{w}_j^l\|_2^2 \label{orthogonal_projection}.
\end{equation}
A higher orthogonality value indicates that the channel is harder to approximate by others, and may have a {better} chance to be retained in the CSS pruning stage of {the} future steps. Thus, the corresponding channel may be sampled with a relatively higher probability. We use the orthogonality values to design our importance sampling distribution, and the probability $p_j^l$ to regrow a channel $\mathbf{w}_j^l$ is given as:
\begin{equation}
     p_j^l=\text{exp}({\epsilon^{l}_j})/\sum\nolimits_{j'}\text{exp}({\epsilon^{l}_{j'}})\label{importance sampling}.
\end{equation}
Then, the channels to regrow are sampled according to the distribution~$\text{Multinomial}(\{p^l_j|j\in[C^l]\setminus\mathcal{T}^l\};\lceil \delta^tC^l \rceil)$ without replacement, where $\delta^t$ is the regrowing factor introduced as follows.

In the regrowing stage, we employ a cosine decay scheduler to gradually reduce the number of regrown channels so that the sub-model converges to the target channel sparsity at the end of training. Specifically, the regrowing factor at $t$-th step is computed as: $\delta_t=\dfrac{1}{2}\left(1+\text{cos}\left(\dfrac{t\cdot\pi}{T_{\text{max}}/\Delta T}\right)\right)\delta_0$, where $\delta_0$ is the initial regrow{ing} factor, $T_{\text{max}}$ denotes the total exploration steps,
and $\Delta T$ represents the frequency to invoke the pruning-regrowing steps. 


\subsection{Sub-model structure exploration}
\label{structure exploration}
{The starting model architecture may not have balanced layer distributions. Some layers are more important to the model accuracy and more channels need to be preserved, while some other layers may contain excessive number of channels.
To better preserve model accuracy, our method dynamically re-distributes the surviving channels across different layers in each pruning stage. Such re-distribution is called sub-model structure exploration.}


{Inspired by \cite{liu2017learning}, we use the learnable scaling factors in batch normalization (BN) \footnote{BN applies affine transformation\f{s} to standardized input feature-maps: $X_{out} = \gamma\tilde{X}_{in} + \beta$, where $\gamma$ / $\beta$ are learnable scaling / shifting factors.} \cite{ioffe2015batch} to reflect the layer importance.
Denote the BN scaling factors of all channels across all layers by $\Gamma=\{\boldsymbol{\gamma}^1,...,\boldsymbol{\gamma}^L\}$, $\boldsymbol{\gamma}^l\in\mathbb R^{C^l}$, and the overall target channel sparsity by $S$. We calculate the layer{-wise} pruning ratios by ranking all scaling factors in descending order and preserving the top $1-S$ percent of the channels. Then, the sparsity $\kappa^l$ for layer $l$ is given as:
\begin{equation}
    \kappa^l=\big({\sum\nolimits_{j\in[C^l]}\mathbbm{1}_{\{\gamma^l_j\leq q(\Gamma,S)\}}}\big) / {C^l},l\in[L], \label{layer-wise sparsity}
\end{equation}
where $\mathbbm{1}_{\{\gamma^l_j\leq q(\Gamma,S)\}}$ is 0 if $\gamma^l_j > q(\Gamma,S)$ and 1 if $\gamma^l_j\leq q(\Gamma,S)$. $q(\Gamma,S)$ represents the $S$-th percentile of all the scaling factors $\Gamma$. Accordingly, the number of channels in each layer of the sub-model is obtained as $\lceil(1-\kappa^l)C^l\rceil$.} 

{\cite{liu2017learning} relies on LASSO regularization to identify \f{the} insignificant channels, it extracts the sub-model from a fully pre-trained model in one-shot manner, and the subsequent finetuning procedure fixes the architecture without adaptation. 
In contrast, our method proposes a CSS-based pruning criterion without requiring any sparse regularization. 
And we advocate a repeated pruning and regrowing paradigm as opposed to the pruning-only strategy.
We use BN scaling factors only for re-allocating the number of channels. We perform such re-allocation repeatedly at each step to take into account the changing layer importance during training.
Thanks to our regrowing stages which help maintain the exploration space, our method can dynamically re-distribute channels from the less crucial layers to the more important ones, leading to a better sub-model structure.}

\section{Theoretical Justification}
\newcommand{\RR}{{\mathbb{R}}}
\newcommand{\EE}{{\mathbb{E}}}
\newcommand{\bW}{{\mathbf{W}}}
\newcommand{\bw}{{\mathbf{w}}}
\newcommand{\cD}{\mathcal{D}}
\newtheorem{proposition}{Proposition}
\newtheorem{assumption}{Assumption}
\newtheorem{remark}{Remark}
\newtheorem*{remark*}{Remark}
\allowdisplaybreaks

We now provide the convergence guarantee for the CHEX method. Let $F(\bW) = \mathbb{E}_{x\sim \cD}[f(\bW;x)]$ be the loss function of the deep learning task where $x$ is the data following a distribution $\cD$ {and $\mathbb{E}[\cdot]$ is the expectation}. In addition, let $\bW_t \in \RR^d$ be the model parameter at the $t$-th training iteration, and $m_t \in \{0,1\}^d$ be a binary \zj{channel} mask 
vector for $t=1, \cdots, T$. Apparently, 
quantity $\bW_t \odot m_t \in \RR^d$ is a sub-model pruned from $\bW_t$ where $\odot$ denotes the element-wise product. The following proposition shows that the CHEX will converge to a neighborhood around the stationary solution at rate $O(1/\sqrt{T})$ when learning rate is set properly. Due to the space limitation, we put its proof in the Appendix.
	
\begin{proposition}[Convergence Guarantee]
	\label{prop-theory}
	Suppose the loss function $F(\bW)$ is $L$-smooth, the sampled stochastic gradient is unbiased and has bounded variance, and the relative error introduced by each mask is bounded, i.e., $\|\bW - \bW \odot m_t\|^2 \le \delta^2 \|\bW\|^2$ and $\|\nabla F(\bW) - \nabla F(\bW) \odot m_t\|^2 \le \zeta^2 \|\nabla F(\bW)\|^2$ for constants $\delta \in [0,1]$ and $\zeta \in [0, 1]$. If learning rate $\eta = \frac{\sqrt{2 C_0}}{\sigma \sqrt{ L (T+1)}}$ in which $C_0 = \EE[F(\bW_0)]$,  the sub-models obtained by CHEX will converge as follows: 
\begin{align}\label{eq-theory}
&\ \frac{1}{T+1}\sum_{t=0}^T \EE[\|\nabla F(\bW_t\odot m_t)\|^2] \nonumber \\
\le&\ \frac{4\sigma\sqrt{L C_0}}{(1 \hspace{-0.5mm}-\hspace{-0.5mm} \zeta)^2\sqrt{T+1}} \hspace{-0.5mm} + \hspace{-0.5mm} \frac{2L^2\delta^2}{(T+1)(1-\zeta)^2} \sum_{t=0}^T \EE [\| \bW_t\|^2].
\end{align}
\end{proposition}

\begin{remark*}
	If there is no pruning (i.e., $m_t = \mathbf{1} \in \RR^d$) in the training process, it holds that $\delta = 0$ and $\zeta = 0$. Substituting it into \eqref{eq-theory}, we find that the CHEX method can converge exactly to the stationary solution, i.e., $\mathbb{E}[\|\nabla F(\bW_T)\|^2] \to 0$ as $T$ increases to infinity.
	When a sparse channel mask 
	is utilized, it holds that $\delta \neq 0$ and $\zeta \neq 0$. In this scenario, the {mask-induced} error will inevitably influence the accuracy of the trained sub-model, i.e., constants $\delta$ and $\zeta$ will influence the magnitude of the second term in the upper bound \eqref{eq-theory}. 
\end{remark*}
%

\section{Experiments}

\begin{table*}[t]
\footnotesize
\centering
\begingroup
\setlength{\tabcolsep}{1pt}
\begin{tabular}[t]{l c c c c || l c c c c}\toprule
    \textbf{Method} & \textbf{PT} & \textbf{FLOPs} & \textbf{Top-1} & \textbf{Epochs} & \textbf{Method} & \textbf{PT} & \textbf{FLOPs} & \textbf{Top-1} & \textbf{Epochs} \\\midrule
    \multicolumn{5}{l||}{\textit{\textbf{ResNet-18}}} & \multicolumn{5}{l}{\textit{\textbf{ResNet-50}}} \\
    \;\;\;PFP \cite{liebenwein2019provable} & Y & 1.27G & 67.4\% & 270 & \;\;\;GBN \cite{you2019gate} & Y & 2.4G & 76.2\% & 350  \\
    \;\;\;SCOP \cite{tang2020scop} & Y & 1.10G & 69.2\% & 230 & \;\;\;LeGR \cite{chin2019legr} & Y & 2.4G & 75.7\% & 150 \\
    \;\;\;SFP \cite{he2018soft} & Y & 1.04G & 67.1\% & 200 & \;\;\;SSS \cite{huang2018data} & N & 2.3G & 71.8\% & 100 \\
    \;\;\;FPGM \cite{he2019filter} & Y & 1.04G & 68.4\% & 200 & \;\;\;TAS \cite{dong2019network} & N & 2.3G & 76.2\% & 240 \\
    \;\;\;DMCP \cite{guo2020dmcp} & N & 1.04G & 69.0\% & 150 & \;\;\;GAL \cite{lin2019towards} & Y & 2.3G & 72.0\% & 150 \\
    \;\;\;\textbf{CHEX} & N &  1.03G & \textbf{69.6\%} & 250 & \;\;\;Hrank \cite{lin2020hrank} & Y & 2.3G & 75.0\% & 570 \\
    \cmidrule{1-5}
    \multicolumn{5}{l||}{\textit{\textbf{ResNet-34}}} & \;\;\;Taylor \cite{molchanov2019importance} & Y & 2.2G & 74.5\% & - \\
    \;\;\;Taylor \cite{molchanov2019importance} & Y & 2.8G & 72.8\% & - & \;\;\;C-SGD \cite{ding2019centripetal} & Y & 2.2G & 74.9\% & - \\ 
    \;\;\;SFP \cite{he2018soft} & Y & 2.2G & 71.8\% & 200 & \;\;\;SCOP \cite{tang2020scop} & Y & 2.2G & 76.0\% & 230 \\
    \;\;\;FPGM \cite{he2019filter} & Y & 2.2G & 72.5\% & 200 & \;\;\;DSA \cite{ning2020dsa} & N & 2.0G & 74.7\% & 120  \\
    \;\;\;GFS \cite{ye2020good} & Y & 2.1G & 72.9\% & 240 & \;\;\;CafeNet \cite{su2021locally} & N & 2.0G & 76.9\% & 300 \\
    \;\;\;DMC \cite{gao2020discrete} & Y & 2.1G & 72.6\% & 490 & \;\;\;\textbf{CHEX-1} & N & 2.0G & \textbf{77.4\%} & 250 \\
    \cmidrule{6-10}
    \;\;\;NPPM \cite{gao2021network} & Y & 2.1G & 73.0\% & 390 & \;\;\;SCP \cite{kang2020operation} & N & 1.9G & 75.3\% & 200 \\
    \;\;\;SCOP \cite{tang2020scop} & Y & 2.0G & 72.6\% & 230 & \;\;\;Hinge \cite{li2020group} & Y & 1.9G & 74.7\% & - \\
    \;\;\;CafeNet \cite{su2021locally} & N & 1.8G & 73.1\% & 300 & \;\;\;AdaptDCP \cite{zhuang2018discrimination} & Y & 1.9G & 75.2\% & 210 \\
    \;\;\;\textbf{CHEX} & N & 2.0G & \textbf{73.5\%} & 250 & \;\;\;LFPC \cite{he2020learning} & Y & 1.6G & 74.5\% & 235 \\
    \cmidrule{1-5}
    \multicolumn{5}{l||}{\textit{\textbf{ResNet-101}}} & \;\;\;ResRep \cite{ding2020lossless} & Y & 1.5G & 75.3\% & 270 \\
    \;\;\;SFP \cite{he2018soft} & Y & 4.4G & 77.5\% & 200 & \;\;\;Polarize \cite{zhuang2020neuron} & Y & 1.2G & 74.2\% & 248 \\
    \;\;\;FPGM \cite{he2019filter} & Y & 4.4G & 77.3\% & 200 & \;\;\;DSNet \cite{li2021dynamic} & Y & 1.2G & 74.6\% & 150 \\
    \;\;\;PFP \cite{liebenwein2019provable} & Y & 4.2G & 76.4\% & 270 & \;\;\;CURL \cite{luo2020neural} & Y & 1.1G & 73.4\% & 190 \\
    \;\;\;AOFP \cite{ding2019approximated} & Y & 3.8G & 76.4\% & - & \;\;\;DMCP \cite{guo2020dmcp} & N & 1.1G & 74.1\% & 150 \\
    \;\;\;NPPM \cite{gao2021network} & Y & 3.5G & 77.8\% & 390 & \;\;\;MetaPrune \cite{liu2019metapruning} & N & 1.0G & 73.4\% & 160 \\
    \;\;\;DMC \cite{gao2020discrete} & Y & 3.3G & 77.4\% & 490 & \;\;\;EagleEye \cite{li2020eagleeye} & Y & 1.0G & 74.2\% & 240  \\
    \;\;\;\textbf{CHEX-1} & N & 3.4G & \textbf{78.8\%} & 250 & \;\;\;CafeNet \cite{su2021locally} & N & 1.0G & 75.3\% & 300 \\
    \;\;\;\textbf{CHEX-2} & N & 1.9G & \textbf{77.6\%} & 250 & \;\;\;\textbf{CHEX-2} & N & 1.0G & \textbf{76.0\%} & 250 \\
    \bottomrule
    \multicolumn{10}{c}{\textbf{(a)}}
\end{tabular}
\endgroup
\hspace{0.11in}
\begingroup
\setlength{\tabcolsep}{0.5pt}
\begin{tabular}[t]{@{}l c c c c c c c @{}}\toprule
    \multirow{2}{*}{\textbf{Method}} & \textbf{FLOPs} & \multirow{2}{*}{\textbf{AP}} & \multirow{2}{*}{\textbf{AP$_{50}$}} & \multirow{2}{*}{\textbf{AP$_{75}$}} & \multirow{2}{*}{\textbf{AP$_S$}} & \multirow{2}{*}{\textbf{AP$_M$}} & \multirow{2}{*}{\textbf{AP$_L$}} \\
     & \textbf{reduction} & & & & & & \\\midrule
    \multicolumn{6}{@{}l}{\textit{\textbf{SSD object detection}}} & \\
    \;\;\;Baseline \cite{liu2016ssd} & 0\% & 25.2 & 42.7 & 25.8 & 7.3 & 27.1 & 40.8 \\
    \;\;\;DMCP \cite{guo2020dmcp} & 50\% & 24.1 & 41.2 & 24.7 & 6.7 & 25.6 & 39.2 \\
    \;\;\;\textbf{CHEX-1} & 50\% & \textbf{25.9} & 43.0 & 26.8 & 7.8 & 27.8 & 41.7 \\
    \;\;\;\textbf{CHEX-2} & 75\% & \textbf{24.3} & 41.0 & 24.9 & 7.1 & 25.6 & 40.1 \\
    \midrule
    \multicolumn{6}{@{}l}{\textit{\textbf{Mask R-CNN object detection}}} & \\
    \;\;\;Baseline \cite{he2017mask} & 0\% & 37.3 & 59.0 & 40.2 & 21.9 & 40.9 & 48.1 \\
    \;\;\;\textbf{CHEX} & 50\% & \textbf{37.3} & 58.5 & 40.4 & 21.7 & 39.0 & 49.5 \\
    \midrule
    \multicolumn{6}{@{}l}{\textit{\textbf{Mask R-CNN instance segmentation}}} & \\
    \;\;\;Baseline \cite{he2017mask} & 0\% & 34.2 & 55.9 & 36.2 & 15.8 & 36.9 & 50.1 \\
    \;\;\;\textbf{CHEX} & 50\% & \textbf{34.5} & 55.7 & 36.7 & 15.9 & 36.1 & 51.2 \\
    \bottomrule
    \multicolumn{8}{c}{\textbf{(b)}} \\
    \vspace{0.225in}
    \\
    \toprule
    \multirow{2}{*}{\textbf{Method}} & \multirow{2}{*}{\textbf{\#Points}} & \multicolumn{2}{c}{\textbf{FLOPs}} & \multicolumn{4}{c}{\multirow{2}{*}{\textbf{Quality}}} \\
    & & \multicolumn{2}{c}{\textbf{reduction}} & \multicolumn{4}{c}{} \\\midrule
    \multicolumn{8}{@{}l}{\textit{\textbf{3D shape classification (Quality is accuracy)}}} \\
    \;\;\;Baseline \cite{qi2017pointnet++} & 1k & \multicolumn{2}{c}{0\%} & \multicolumn{4}{c}{92.8\%} \\
    \;\;\;\textbf{CHEX} & 1k & \multicolumn{2}{c}{87\%} & \multicolumn{4}{c}{\textbf{92.9\%}} \\\midrule
    \multicolumn{8}{@{}l}{\textit{\textbf{3D part segmentation (Quality is class/instance mIoU)}}} \\
    \;\;\;Baseline \cite{qi2017pointnet++} & {2k} & \multicolumn{2}{c}{0\%} & \multicolumn{4}{c}{82.5\%/85.4\%} \\
    \;\;\;ADMM \cite{ren2019admm} & 2k & \multicolumn{2}{c}{90\%} & \multicolumn{4}{c}{77.1\%/84.0\%} \\
    \;\;\;\textbf{CHEX} & 2k & \multicolumn{2}{c}{91\%} & \multicolumn{4}{c}{\textbf{82.3\%/85.2\%}} \\
    \bottomrule
    \multicolumn{8}{c}{\textbf{(c)}}
\end{tabular}
\endgroup
\caption{\textbf{(a)}: Results of ResNet on ImageNet dataset. ``PT'': require pre-training. ``Y'': Yes, ``N'': No. \textbf{(b)}: Results of SSD on COCO2017 and Mask R-CNN on COCO2014. For objection detection, we evaluate the bounding box AP. For instance segmentation, we evaluate the mask AP. \textbf{(c)}: Results of PointNet++ for 3D point clouds classification on ModelNet40 dataset and segmentation on ShapeNet dataset. 
}
\label{tab:main_results}
\vspace{-0.1in}
\end{table*}


To {evaluate} the efficacy and generality of CHEX, we {experiment} on a variety of computer vision tasks with diverse CNN architectures.
All experiments run {on PyTorch framework} {based on DeepLearningExamples \cite{nvidia_github}} with NVIDIA Tesla V100 GPUs.
We set $\delta_0=0.3$ and $\Delta T=2~\text{epoch}$, where $\delta_0$ means the initial regrowing {factor}, and $\Delta T$ is the number of training iterations between two consecutive pruning-regrowing steps.
To keep our method simple and generic, the above hyper-parameters are kept constant for our experiments. 
{We set the rest of the hyper-parameters in the} default settings and specify them in the Appendix. {In this paper, FLOPs is calculated by counting multiplication and addition as one operation by following~\cite{he2016deep}}.

\subsection{Image recognition}

For image recognition, 
we apply CHEX to ResNet \cite{he2016deep} with different depths
on ImageNet \cite{imagenet_cvpr09}. 
The baseline ResNet-18/34/50/101 models have 1.8/3.7/4.1/7.6 GFLOPs with 
70.3\%/73.9\%/77.8\%/78.9\% top-1 accuracy, respectively.

As shown in Table~\ref{tab:main_results}(a), CHEX achieves noticeably higher accuracy than {the} state-of-the-art channel pruning methods under the same FLOPs. For example, compared with MetaPruning \cite{liu2019metapruning}, DCMP \cite{guo2020dmcp} and CafeNet~\cite{su2021locally}, our pruned ResNet-50 {model} with 4$\times$ FLOPs reduction achieve{s} 2.6\%, 1.6\%, and 0.7\% higher top-1 accuracy, respectively. 
On the other hand, at the same target accuracy, CHEX achieve{s} higher FLOPs reduction. For example, CHEX achieves 4$\times$ FLOPs reduction on ResNet-101 {model} with 77.6\% top-1 accuracy,
compared to the latest work NPPM \cite{gao2021network} which yields 2.2$\times$ FLOPs reduction. 

The results in Table~\ref{tab:main_results}(a) also show an interesting property of our CHEX method. When we search a sub-{model} with a small FLOPs target, it is better to {start} our method on a larger model than on a smaller one. For instance, pruning {a} ResNet-50 {model} to 1~GFLOPs yields an accuracy 6.6\% higher than pruning {a} ResNet-18 {model} to 1~GFLOPs. This {indicates} {that} CHEX performs training-time structural exploration more effectively when given a larger parametric space. 
To illustrate this point further, we conduct an additional experiment by
{applying CHEX to a model with twice the number of channels as the ResNet-50 model, with the goal of {reducing its FLOPs to the same level} as the original ResNet-50 model.}
Notably, {this sub-model} achieves 78.9\% top-1 accuracy at 4.1 GFLOPs. This suggests that CHEX has the potential to 
{optimize} existing CNNs.

\subsection{Object detection}

For object detection, we apply CHEX to \f{the} SSD \f{model}~\cite{liu2016ssd} on COCO2017 dataset \cite{lin2014microsoft}. 
Table~\ref{tab:main_results}(b) summarizes the performance with different FLOPs reductions. Our pruned model outperforms the baseline model by 0.7\% AP (25.9 vs. 25.2) while achieving 2$\times$ FLOPs reduction.
Compared with previous SOTA channel pruning method \cite{guo2020dmcp}, our method achieves 1.8\% higher AP (25.9 vs. 24.1) with 2$\times$ FLOPs reduction. 
Moreover, our method achieves 4$\times$ FLOPs reduction with less than 1\% AP loss.

\subsection{Instance segmentation}

For instance segmentation, we apply CHEX to the Mask R-CNN \f{model~}\cite{he2017mask} on COCO2014 dataset. 
Since Mask R-CNN is a multi-task framework, we evaluate both the bounding box AP for object detection and the mask AP for instance segmentation. As shown in Table \ref{tab:main_results}(b),
{the models pruned using} our method achieve 2$\times$ FLOPs reduction without AP loss, even for the challenging instance segmentation task where the model needs to detect all objects {correctly} in an image while precisely segmenting each instance.

\subsection{3D point cloud}
Apart from 2D computer vision problems, we also 
apply CHEX to compress PointNet++ \cite{qi2017pointnet++} for 3D shape classification on ModelNet40 \cite{wu20153d} dataset and 3D part segmentation on ShapeNet \cite{yi2016scalable} dataset. 
The results are shown in Table \ref{tab:main_results}(c).
On shape classification, {the model pruned using} our method achieves around 7.5$\times$ FLOPs reduction while improving the accuracy slightly compared to the unpruned baseline. On part segmentation, {the model pruned using} our method achieves 11$\times$ FLOPs reduction while maintaining similar mIoU compared with the unpruned baseline.

\section{Ablation Analysis}

We investigate {the} effectiveness of different components in the CHEX method through ablation studies. All {the} following results are based on pruning the ResNet-50 {model} to 1~GFLOPs (4$\times$ reduction) on ImageNet {dataset}.

\begin{table}[t]
\centering
\footnotesize
\begin{tabular}{@{}l c@{}}\toprule
    \textbf{Method} & \textbf{Top-1}  \\\midrule
    \;Baseline & 73.9\% \\
    \;\;\;+ Periodic pruning and regrowing & 74.8\% \\
    \;\;\;+ Dynamic sub-model structure exploration & 75.6\% \\
    \;\;\;+ Importance sampling-based regrowing & 76.0\% \\
    \bottomrule
\end{tabular}
\caption{Ablation study of different components in CHEX.}
\label{tab:ablation}
\end{table}

\noindent\textbf{Ablation study.}
In Table \ref{tab:ablation}, we study the effectiveness of different components in CHEX, namely the periodic channel pruning and regrowing process, {the} sub-model structure exploration technique,
and the importance sampling-based regrowing.
The baseline is one-shot early-stage pruning~\cite{you2019drawing}, where the model is pruned by CSS very early after 6\% of the total training epochs, and there is no regrowing {stage}. The sub-model architecture is based on the BN scaling factors and kept fixed in \f{the} subsequent training.
The periodic pruning and regrowing process 
{repeatedly samples the important channels and prevents \f{the} channels from being pruned prematurely.} This brings in $0.9\%$ accuracy improvement. 
When the number of channels in each layer is also dynamically adjusted instead of sticking to the fixed structure determined very early in training, we can obtain \f{a} more optimal sub-model structure, which \f{further improves the accuracy by $0.8\%$}.
Finally, using the importance sampling strategy \f{described in} Eq.\eqref{importance sampling} instead of uniform sampling in the regrowing stages improves the top-1 accuracy to 76\%.

\noindent\textbf{Influence of pruning criterion.}
We study the influence of pruning criterion to the final accuracy by comparing CSS versus filter magnitude \cite{li2016pruning} and BN \cite{liu2017learning} pruning in Table \ref{tab:prune_criterion}. 
{As suggested by \cite{li2016pruning,luo2017thinet,molchanov2019importance,hou2020feature}, pruning criterion is an important factor in the traditional pretrain-prune-finetune (PPF) approach. Indeed, when the PPF approach is used, our proposed CSS shows the best accuracy, outperforming magnitude and BN pruning by $0.8\%$ and $1.5\%$, respectively.}
On the other hand, we observe that CHEX is more robust to the pruning criterion, as it improves the accuracy for all three criteria and the accuracy gap among them becomes smaller: CSS outperforms magnitude and BN pruning by $0.3\%$ and $0.7\%$, respectively.
We suspect that this is because CHEX can dynamically {adjust the channel importance} and can recover from sub-optimal pruning decisions. 


\begingroup
\setlength{\tabcolsep}{6pt}
\begin{table}[t]
\centering
\footnotesize
\begin{tabular}{@{}l c c c@{}} \toprule
    {\textbf{Prune criterion}} & \textbf{BN} \cite{liu2017learning} & \textbf{Magnitude} \cite{li2016pruning} & \textbf{CSS (Ours)} \\\midrule
    Pretrain-prune-finetune & 73.1\% & 73.8\% & 74.6\% \\\midrule
    {CHEX} & 75.3\% & 75.7\% & 76.0\%  \\\bottomrule
\end{tabular}
\vspace{-0.1in}
\caption{Influence of \f{the} pruning criterion to the top-1 accuracy in pretrain-prune-finetune framework and our CHEX method.}
\label{tab:prune_criterion}
\end{table}
\endgroup

\noindent\textbf{Design choices for the regrowing stage.}
In Table \ref{tab:regrowing_ablation}, we investigate the impact of different schemes in the channel regrowing stage of CHEX:

\begingroup
\setlength{\tabcolsep}{2.7pt}
\begin{table}[t]
\centering
\footnotesize
\begin{tabular}{@{}l c c c c c@{}}\toprule
    \multicolumn{5}{@{}l}{\textit{\textbf{Initialization of regrown channels}}} \\
    \;\;{Design choices} & Zero & Random & EMA & \multicolumn{2}{c}{MRU} \\
    \;\;Top-1 & 74.1\% & 74.5\% & 75.5\% & \multicolumn{2}{c}{76.0\%} \\
    \toprule
    \multicolumn{5}{@{}l}{\textit{\textbf{Regrowing factor}}} \\
    \;\;Design choices & $\delta_0=0.1$ & $\delta_0=0.2$ & $\delta_0=0.3$ & $\delta_0=0.4$ & Full \\
    \;\;Top-1 & 75.2\% & 75.6\% & 76.0\% & 75.9\% & 76.0\% \\
    \toprule
    \multicolumn{5}{@{}l}{\textit{\textbf{Scheduler for channel regrowing}}} \\
    \;\;Design choices & Constant & \multicolumn{2}{c}{Linear decay} & \multicolumn{2}{c}{Cosine decay} \\
    \;\;Top-1 & {75.3\%} & \multicolumn{2}{c}{75.6\%} & \multicolumn{2}{c}{76.0} \\
    \toprule
    \multicolumn{5}{@{}l}{\textit{\textbf{Pruning-regrowing frequency}}} \\
    \;\;Design choices & $\Delta T=1$ & $\Delta T=2$ & $\Delta T=5$ & $\Delta T=10$ & $\Delta T=20$ \\
    \;\;Top-1 & 75.2\% & 76.0\% & 75.8\% & 75.3\% & 74.9\% \\
    \bottomrule
\end{tabular}
\vspace{-0.1in}
\caption{Compare different design choices \f{in} the regrowing stages of the CHEX method.}
\label{tab:regrowing_ablation}
\end{table}
\endgroup

(1) Different initialization schemes for the regrown channels affect model quality critically. 
{We have experimented several methods by initializing the regrown channels with random normal distribution \cite{he2015delving}, zero weights, the exponential moving average (EMA) weights, and most recently used (MRU) weights.}
The results show that MRU yields the best top-1 accuracy, outperforming the other three initialization schemes by $0.5\% \sim 1.9\%$.

(2) The initial regrowing factor $\delta_0$ also affects the model quality.
Intuitively, a large regrowing factor can maintain relatively larger model capacity in the early stage of training, and involve more channels into the exploration steps. As shown, the accuracy {is} improved by 0.8\% as the $\delta_0$ increases from $0.1$ to $0.3$, at {the} price of {more} training {cost}. However, we did not observe further improvement when the regrowing factor increases from $\delta_0=0.3$ to full model size, in which case one step in CHEX consists of pruning to target channel sparsity and regrowing all pruned channels. This suggests that regrowing \textit{a fraction} of \f{the} previously pruned channels may be enough for a comprehensive exploration of the channels if we also aim more training cost savings.

(3) We decay the number of regrown channels in each regrowing stage to gradually restrict the scope of exploration. We compare several decay schedulers, including constant, linear, and cosine. The cosine decay scheduler performs better than the other two schemes by 0.7\% and 0.4\% accuracy, respectively. With a decay scheduler, the sub-model under training will enjoy gradually decreased computation cost.

(4) {The training iterations between two consecutive pruning-regrowing steps, $\Delta T$, also affects the model quality. 
A smaller $\Delta T$ incurs more frequent pruning-regrowing steps in CHEX, leading to better accuracy in general. We use $\Delta T=2$~epoch, as it gives the best accuracy.}

\vspace{-0.1in}
\section{Related Works}
\vspace{-0.1in}

\noindent\textbf{Channel pruning} methods can be roughly categorized into three classes. 

(1) \textit{Pruning after training} approaches follow {a} three-stage pipeline: pre-training a large model, pruning the unimportant channels, and finetuning the pruned model.
Prior arts in this category mainly {focus on comparing} different pruning criteria. Exemplary metrics for evaluating channel importance include weight norm~\citep{li2016pruning,he2018soft,yang2018netadapt,chin2019legr,he2020learning}, geometric median~\citep{he2019filter}, Taylor expansion of cross-entropy loss~\cite{lin2018accelerating,you2019gate,molchanov2019importance}, discrepancy of final response layer~\citep{yu2018nisp}, feature-maps reconstruction error~\citep{he2017channel,luo2017thinet,zhuang2018discrimination}, feature-maps rank~\citep{lin2020hrank}, KL-divergence~\cite{luo2020neural}, greedy forward selection with largest loss reduction~\cite{ye2020good}, feature-maps discriminant information \cite{hou2020feature,hou2020efficient,kung2020augment}.
Our method differs substantially from these approaches that we do not pre-train a large model.
{Instead, the compact sub-model is explored during a normal training process from scratch.}

(2) \textit{Pruning during training} approaches~\cite{liu2017learning,huang2018data,lin2019towards,li2019compressing,guo2020multi,zhuang2020neuron,li2020group,kang2020operation,ye2018rethinking,guo2016dynamic,hou2022multi,yuan2021mest,ma2021effective} perform channel selection and model training jointly, usually by imposing sparse regularization{s} to the channel-wise scaling factors and adopting specialized optimizers to solve it.
Although these methods usually yield good acceleration due to joint optimization,
many of them~\cite{liu2017learning,lin2019towards,li2019compressing,guo2020multi,zhuang2020neuron,li2020group,zhang2021unified} perform the sparse regularization training on fully pre-trained large models. Therefore, these approaches still suffer from the expensive training cost. {In contrast}, our method does not rely on {sparse} regularization. Instead, the periodic channel pruning and regrowing processes adaptively explore the channel importance during training.

(3) \textit{Pruning at early stage} methods compress the model at random initialization or early stage of training~\cite{you2019drawing,wang2019pruning,lee2018snip,wang2020picking,tanaka2020pruning, liu2018rethinking,frankle2018lottery,ma2021sanity}.
Although the training cost is reduced, one-shot pruning based on under-trained model weights may not properly reflect the channel importance and the model representation ability is prematurely reduced at the very beginning, {resulting in} significant accuracy loss.
In contrast, the proposed {method} can recover prematurely pruned channels and better maintain the model capacity.

\noindent\textbf{AutoML or NAS based pruning} {methods automatically search} the optimal number of channels in each layer of CNN.
\cite{he2018amc} uses reinforcement learning to search a compression policy in a layer-wise manner. MetaPruning \cite{liu2019metapruning} pre-trains a hyper-network to predict the weights for candidate pruned models, and adopts evolutionary search to find a good candidate.
NetAdapt \cite{yang2018netadapt} progressively adapts a pre-trained model to a mobile platform until a resource budget is met.
DMCP \cite{guo2020channel} proposes a differentiable search method by modeling channel pruning as a Markov process. 
CafeNet \cite{su2021locally} proposes locally free weight sharing for the one-shot model width search.
\cite{hou2021multi} proposes a Gaussian process search for optimizing the multi-dimensional compression policy.
These approaches usually rely on a pre-trained supernet, and require {a} separate search process involving substantial search-evaluation iterations.
In contrast, CHEX performs sub-{model} structure learning together with weights optimization, making it more efficient than the search-based methods.
Channel pruning has also been incorporated into NAS to further tune the searched architecture for different latency targets \cite{cai2019once,howard2019searching}.
Since CHEX
does not increase the training time, it can be potentially used {as a step in AutoML.}



\vspace{-0.1in}
\section{Conclusion}
\vspace{-0.1in}

We propose a novel channel exploration methodology, CHEX, to reduce the computation cost of deep CNNs in both training and inference. CHEX (1) dynamically adjusts the channel importance based on a periodic pruning and regrowing process, which prevents the important channels from being prematurely pruned; (2) dynamically re-allocates the number of channels under a global sparsity constraint to search for {the} optimal sub-model structure during training.
We design the components in the pruning and regrowing process by proposing a column subset selection based criterion for pruning and a channel orthogonality based importance sampling for regrowing. {This enables us to} obtain a sub-model with high accuracy in only one training pass from scratch, without pre-training a large model or requiring extra fine-tuning.
Experiments on multiple deep learning tasks demonstrate that our method can effectively reduce the FLOPs of diverse CNN models while achieving superior accuracy compared to the state-of-the-art methods. 

\vspace{-0.2in}
\paragraph{Acknowledgment}
This work was supported by Alibaba Group through Alibaba Research Intern Program.

{\small
\bibliographystyle{ieee_fullname}
\bibliography{egbib}
}

\clearpage
\section*{Appendix}

\section*{A1.~~~Implementation details}

\noindent\textbf{Image classification on ImageNet}.
To train ResNet with \f{the} CHEX method on ImageNet {dataset}, we use SGD optimizer with a momentum of 0.875, a mini-batch size of 1024\f{,} and an initial learning rate of 1.024. The learning rate is linearly warmed up for the first 8 epochs, and decayed to zero by a cosine learning rate schedule.
The weight decay is set to 3e-5. Same as previous methods \cite{guo2020dmcp,liu2019metapruning,guo2021gdp}, we also use label smoothing with factor 0.1.
We train the model for a total of 250 epochs.
For data augmentation, we only use random resized crop to 224$\times$224 resolution, random horizontal flip, and normalization. 

\vspace{0.07in}
\noindent\textbf{Object detection on COCO2017}.
Following \cite{liu2016ssd}, we train SSD with the CHEX method on COCO train2017 split containing about 118k images and evaluate on the val2017 split containing 5k images. The input size is fixed to $300\times300$. 
We adopt SGD optimizer with a momentum of 0.9, a mini-batch size of 64, and a weight decay of 5e-4. We train the model for a total of 240k iterations. The initial learning rate is set to 1e-3, and is decayed by 10 at the 160k and 200k iteration.
The SSD uses {a} ResNet-50 {model pretrained on ImageNet dataset} as the backbone.

\vspace{0.07in}
\noindent\textbf{Instance segmentation on COCO2014}.
We follow the standard practice as \cite{he2017mask} to train Mask R-CNN with the CHEX method on COCO training split, and evaluate on the validation split. 
We train with a batch size of 32 for 160K iterations. {We adopt} SGD with {a} momentum {of} 0.9 and a weight decay of 1e-4. The initial learning rate is set to 0.04, which is decreased by 10 at the 100k and 140k iteration. The Mask R-CNN uses ResNet-50-FPN {model} as the backbone.

\vspace{0.07in}
\noindent\textbf{3D classification and segmentation on ModelNet40 and ShapeNet}.
Following \cite{qi2017pointnet++}, we train the PointNet++ model with the CHEX method using the
Adam optimizer with a mini-batch size of 32. The learning rate begins with 0.001 and decays with a rate of 0.7 every 20 epochs.  We train the model for a total of 200 epochs on the ModelNet40 for 3D shape classification, and 250 epcohs on the ShapeNet dataset for 3D part segmentation.


\section*{A2.~~~More results}

\subsection*{A2.1.~~~CHEX on lightweight CNNs}
\label{sec:lightweight_cnns}
We apply the CHEX method to compress compact CNN models MobileNetV2 and EfficientNet-B0. As shown in \Cref{tab:lightweight_cnns}, our compressed MobileNetV2 model with around 30\% FLOPs reduction achieves almost no accuracy loss compared to the unpruned baseline. With 50\% FLOPs reduction, our compressed MobileNetV2 model outperforms previous state-of-the-art channel pruning methods by 0.8$\sim$2.3\% accuracy. Similarly, our method achieves superior accuracy when compressing EfficientNet-B0 with the same FLOPs reduction as the previous methods.

\begin{table}[h]
\centering
\footnotesize
\begin{tabular}[t]{l l c c}\toprule
    \textbf{Model} & \textbf{Method} & \textbf{FLOPs} & \textbf{Top-1}  \\\midrule
    \multirow{15}{*}{\;MobileNetV2} & \;Baseline & 300M & 72.2\% \\ 
    & \;LeGR \cite{chin2019legr} & 220M & 71.4\% \\
    & \;GFS \cite{ye2020good} & 220M & 71.6\% \\ 
    & \;MetaPruning \cite{liu2019metapruning} & 217M & 71.2\% \\
    & \;DMCP \cite{guo2020dmcp} & 211M & 71.6\% \\
    & \;AMC \cite{he2018amc} & 210M & 70.8\% \\
    & \;PFS \cite{wang2019pruning} & 210M & 70.9\% \\
    & \;JointPruning \cite{liu2020joint} & 206M & 70.7\% \\
    & \;\textbf{CHEX-1} & 220M & \textbf{72.0\%} \\\cmidrule{2-4}
    & \;DMC \cite{gao2020discrete} & 162M & 68.4\% \\ 
    & \;GFS \cite{ye2020good} & 152M & 69.7\% \\
    & \;LeGR \cite{chin2019legr} & 150M & 69.4\% \\
    & \;JointPruning \cite{liu2020joint} & 145M & 69.1\% \\
    & \;MetaPruning \cite{liu2019metapruning} & 140M & 68.2\% \\
    & \;\textbf{CHEX-2} & 150M & \textbf{70.5\%} \\ \midrule
    \multirow{7}{*}{\;EfficientNet-B0} & \;Baseline & 390M & 77.1\% \\
    & \;PEEL \cite{hou2021network} & 346M & 77.0\% \\
    & \;\textbf{CHEX-1} & 330M & \textbf{77.4\%} \\\cmidrule{2-4}
    & \;DSNet \cite{li2021dynamic} & 270M & 75.4\% \\
    & \;\textbf{CHEX-2} & 270M & \textbf{76.2\%} \\\cmidrule{2-4}
    & \;CafeNet-R \cite{su2021locally} & 192M & 74.5\% \\
    & \;\textbf{CHEX-3} & 192M & \textbf{74.8\%} \\
    \bottomrule
\end{tabular}
\caption{Results of MobileNetV2 and EfficientNet-B0 on ImageNet dataset.}
\label{tab:lightweight_cnns}
\end{table}

\subsection*{A2.2.~~~Comparison with GrowEfficient and PruneTrain}
We follow GrowEfficient's settings\f{~\cite{YuanSM21}} in choosing WideResNet-28-10 (on CIFAR10 dataset) and ResNet-50 (on ImageNet dataset) as the baseline model{s}. For a fair comparison, we adopt the same training hyper-parameters as GrowEfficient, and train the models with CHEX from scratch.
Both GrowEfficient and PruneTrain \cite{lym2019prunetrain} sparsify the models using LASSO regularization during training.
In contrast, the CHEX method incorporates explicit channel pruning and regrowing stages, and interleaves them in a repeated manner without any sparse regularization.
As shown {in Table~\ref{tab:compare_growefficient}}, our method achieves noticeably higher accuracy than GrowEfficient and PruneTrain under same FLOPs reduction. 
Moreover, our method demonstrates effective training cost saving compared to the baseline model training without accuracy loss.

\begin{table}[h]
\centering
\footnotesize
\begin{tabular}{@{}l c c c c@{}} \toprule
    \multirow{2}{*}{\textbf{Method}} & \textbf{FLOPs} & \multirow{2}{*}{\textbf{Top-1}} &  \textbf{Training} \\
    &  \textbf{reduction} & &  \textbf{cost saving} \\\midrule
    \multicolumn{4}{@{}l}{\textit{\textbf{WRN-28-10 on CIFAR10 (200 epochs)}}} \\
    \;\;\;Baseline \cite{YuanSM21} & 0\% & 96.2\% & 0\% \\
    \;\;\;GrowEfficient \cite{YuanSM21} & 71.8\% & 95.3\% & 67.9\% \\
    \;\;\;\textbf{CHEX} & 74.8\% & \textbf{96.2\%} & 48.8\% \\\midrule
    \multicolumn{4}{@{}l}{\textit{\textbf{ResNet-50 on ImageNet (100 epochs)}}} \\
    \;\;\;Baseline \cite{YuanSM21} & 0\% & 76.2\% & 0\% \\
    \;\;\;PruneTrain \cite{lym2019prunetrain} & 44.0\% & 75.0\% & 30.0\% \\
    \;\;\;GrowEfficient \cite{YuanSM21} & 49.5\% & 75.2\% & 47.4\% \\
    \;\;\;\textbf{CHEX} & 50.2\% & \textbf{76.3\%} & 43.0\% \\
    \bottomrule
\end{tabular}
\caption{Comparison with GrowEfficient and PruneTrain on CIFAR10 and ImageNet datasets. All methods train from scratch with the same number of epochs. 
}
\label{tab:compare_growefficient}
\end{table}

\subsection*{A2.3.~~~Comparison with NAS}
We also compare the CHEX method with the state-of-the-art NAS method, OFA \cite{cai2019once} on ImageNet. 
For a fair comparison, we take ResNet-50D \cite{he2019bag} as the baseline architecture to perform our CHEX method, by following OFA-ResNet-50~\cite{OFA_github}. 
OFA firstly trains a supernet, then applies progressive shrinking in four dimensions, including number of layers, number of channels, kernel sizes, and input resolutions, and finally finetunes the obtained sub-models. 
In contrast, our CHEX method only adjusts the number of channels via the periodic pruning and regrowing process, and we do not require training a supernet nor extra finetuning.
As shown in \Cref{tab:compare_ofa}, CHEX achieves superior accuracy under similar FLOPs constraints but with significantly less model parameters and training GPU hours than OFA.

\begin{table}[h]
\centering
\footnotesize
\begin{tabular}{@{}l c c c c@{}}\toprule
    \multirow{2}{*}{\textbf{Method}} & \multirow{2}{*}{\textbf{FLOPs}} & \multirow{2}{*}{\textbf{Params.}} & \multirow{2}{*}{\textbf{Top-1}} & \textbf{Training cost} \\
    & & & & \textbf{(GPU hours)} \\\midrule
    \;OFA \cite{cai2019once,OFA_github} & 900M & 14.5M & 76.0\% & 1200 \\
    \;OFA\#25 \cite{cai2019once, OFA_github} & 900M & 14.5M & 76.3\% & 1200 \\
    \;\textbf{CHEX} & 980M & 7.2M & \textbf{76.4\%} & 130 \\
    \;\textbf{CHEX}$_{2\times}$ & 980M & 7.2M & \textbf{76.8\%} & 260 \\\bottomrule
\end{tabular}
\caption{Comparison with OFA using ResNet-50D {model} on ImageNet {dataset}. ``CHEX$_{2\times}$'' means doubling the training epochs in our method.}
\label{tab:compare_ofa}
\end{table}

\subsection*{A2.4.~~~CHEX from pretrained models}
To further showcase the generality of our method, we apply CHEX to a pretrained model. For a fair comparison with other pretrain-prune-finetune methods, we use the pretrained ResNet models provided by the \dl{t}\f{T}orchvision model zoo \footnote{\url{https://pytorch.org/vision/stable/models.html}}. In this setup, CHEX runs for 120 training epochs to match the finetuing epochs of most of the previous methods. As shown in \Cref{tab:gap_from_pretrained_model}, our method achieves competitive \f{top}\dl{Top}-1 accuracy when reducing the same amount of FLOPs compared to previous state-of-the-art pretrain-prune-finetune methods.

\begin{table}[h]
\footnotesize
\centering
\begin{tabular}[t]{l l c c c}\toprule
    \textbf{Model} & \textbf{Method} & \textbf{FLOPs} & \textbf{Top-1} & \textbf{Epochs} \\\midrule
    \multirow{6}{*}{\;ResNet-18} 
    & \;Baseline & 1.81G & 69.4\% & 90 \\
    & \;PFP \cite{liebenwein2019provable} & 1.27G & 67.4\% & 90+180 \\
    & \;SCOP \cite{tang2020scop} & 1.10G & 69.2\% & 90+140 \\
    & \;SFP \cite{he2018soft} & 1.04G & 67.1\% & 100+100 \\
    & \;FPGM \cite{he2019filter} & 1.04G & 68.4\% & 100+100 \\
    & \;\textbf{CHEX} &  1.04G & {69.2\%} & 90+120 \\
    \midrule
    \multirow{8}{*}{\;ResNet-34} 
    & \;Baseline & 3.7G & 73.3\% & 90 \\
    & \;SFP \cite{he2018soft} & 2.2G & 71.8\% & 100+100 \\ 
    & \;FPGM \cite{he2019filter} & 2.2G & 72.5\% & 100+100 \\
    & \;GFS \cite{ye2020good} & 2.1G & 72.9\% & 90+150 \\
    & \;DMC \cite{gao2020discrete} & 2.1G & 72.6\% & 90+400 \\
    & \;NPPM \cite{gao2021network} & 2.1G & 73.0\% & 90+300 \\
    & \;SCOP \cite{tang2020scop} & 2.0G & 72.6\% & 90+140 \\
    & \;\textbf{CHEX} & 2.0G & {72.7\%} & 90+120 \\
    \midrule
    \multirow{14}{*}{\;ResNet-50} 
    & \;Baseline & 4.1G & 76.2\% & 90 \\
    & \;SFP \cite{he2018soft} &  2.4G & 74.6\% & 100+100 \\
    & \;FPGM \cite{he2019filter} & 2.4G & 75.6\% & 100+100 \\
    & \;GBN \cite{you2019gate} & 2.4G & 76.2\% & 90+260 \\
    & \;LeGR \cite{chin2019legr} & 2.4G & 75.7\% & 90+60 \\
    & \;GAL \cite{lin2019towards} & 2.3G & 72.0\% & 90+60 \\
    & \;Hrank \cite{lin2020hrank} & 2.3G & 75.0\% & 90+480 \\
    & \;SRR-GR \cite{wang2021convolutional} & 2.3G & 75.8\% & 90+150 \\
    & \;Taylor \cite{molchanov2019importance} & 2.2G & 74.5\% & 90+25 \\
    & \;C-SGD \cite{ding2019centripetal} & 2.2G & 74.9\% & - \\
    & \;SCOP \cite{tang2020scop} & 2.2G & 76.0\% & 90+140 \\
    & \;DSNet \cite{li2021dynamic} & 2.2G & 76.1\% & 150+10 \\
    & \;EagleEye \cite{li2020eagleeye} & 2.0G & 76.4\% & 120+120 \\
    & \;\textbf{CHEX} & 2.0G & {76.8\%} & 90+120 \\
    \midrule
    \multirow{8}{*}{\;ResNet-101}
    & \;Baseline & 7.6G & 77.4\% & 90 \\
    & \;SFP \cite{he2018soft} & 4.4G & 77.5\% & 100+100 \\
    & \;FPGM \cite{he2019filter} & 4.4G & 77.3\% & 100+100 \\
    & \;PFP \cite{liebenwein2019provable} & 4.2G & 76.4\% & 90+180 \\
    & \;AOFP \cite{ding2019approximated} & 3.8G & 76.4\% & - \\
    & \;NPPM \cite{gao2021network} & 3.5G & 77.8\% & 90+300 \\
    & \;DMC \cite{gao2020discrete} & 3.3G & 77.4\% & 90+400 \\
    & \;\textbf{CHEX} & 3.0G & 78.2\% & 90+120 \\\bottomrule
\end{tabular}
\caption{\f{Compress ResNets starting from the pretrained models. All models are trained on the ImageNet dataset.} ``Epochs'' are reported as: pretraining epochs \f{plus} all subsequent training epochs needed to obtain the final pruned model.}
\label{tab:gap_from_pretrained_model}
\end{table}



\subsection*{A2.5.~~~Comparison with gradual pruning}
{
To further evidence the necessity of the channel exploration via the repeated pruning-and-regrowing approach, we compare CHEX with gradual pruning, where the channel exploration is changed to an iterative pruning-training process with gradually increased channel sparsity (but without regrowing).
For a fair comparison, we apply the CSS pruning criterion, determine the number of channels in each layer by the batch-norm scaling factors, and use the single training pass from scratch when perform gradual pruning. As shown in Table \ref{tab:chex_vs_gradual}, CHEX outperforms gradual pruning by 1.1\% accuracy under the same training setup.
}

\begin{table}[h]
\centering
\footnotesize
\begin{tabular}{@{}l c c@{}}\toprule
    \textbf{Method} & \textbf{FLOPs} & \textbf{Top-1} \\\midrule
    \;Gradual pruning & 1.0G & 74.9\% \\
    \;\textbf{CHEX} & 1.0G & \textbf{76.0\%} \\
    \bottomrule
\end{tabular}
\caption{Comparison with gradual pruning. Results are based on pruning ResNet-50 by 75\% FLOPs on ImageNet.}
\label{tab:chex_vs_gradual}
\end{table}

\subsection*{A2.6.~~~Prune models with shortcut connections}
{We have experimented two strategies to deal with the shortcut connections: (1) Prune
internal layers (e.g., the first two convolution layers in the bottleneck
blocks of ResNet-50), leaving the layers with residual connections
unpruned as \cite{zhuang2018discrimination,luo2017thinet}; (2) Use group pruning as \cite{you2019gate}, where the channels connected by the shortcut connections are pruned simultaneously by summing up their CSS scores. As shown in Table \ref{tab:shortcut}, the first strategy gave better accuracy at less FLOPs, thus we adopted the first strategy in our CHEX method when pruning models with shortcut connections.}

\begin{table}[h]
\centering
\footnotesize
\begin{tabular}{@{}l c c@{}}\toprule
    \textbf{Method} & \textbf{FLOPs} & \textbf{Top-1} \\\midrule
    \;Group pruning & 1.4G & 75.9\% \\
    \;Prune internal layers & 1.0G & 76.0\% \\
    \bottomrule
\end{tabular}
\caption{Comparison of two strategies for pruning models with shortcut connections, using ResNet-50 on ImageNet as an example.}
\label{tab:shortcut}
\end{table}

\subsection*{A2.7.~~~Probabilistic sampling vs. deterministic selection for channel regrowing}
{
In channel regrowing stages, the channels to regrow are sampled according to a probabilistic distribution derived from the channel orthogonality. 
The importance sampling encourages exploration in the regrowing stages, as the randomness gives a chance to sample channels other than the most orthogonal ones. 
Our ablation in Table \ref{tab:sampling_vs_deterministic} shows that the probabilistic sampling achieves better accuracy. than deterministic selection based on channel orthogonality.}

\begin{table}[h]
\centering
\footnotesize
\begin{tabular}{@{}l c c@{}}\toprule
    \textbf{Method} & \textbf{FLOPs} & \textbf{Top-1} \\\midrule
    \;Deterministic selection & 1.0G & 75.7\% \\
    \;Probabilistic sampling & 1.0G & 76.0\% \\
    \bottomrule
\end{tabular}
\caption{Comparison of two schemes for determining the channels to regrow. Results are based on pruning ResNet-50 by 75\% FLOPs on ImageNet.}
\label{tab:sampling_vs_deterministic}
\end{table}

\subsection*{A2.8.~~~Actual inference runtime acceleration}
{In the main paper, we chose FLOPs as our evaluation criteria in order to compare with the prior arts, the majority of which evaluate in terms of FLOPs and accuracy.
Meanwhile, we have compared the actual inference throughput (images per second) of our compressed models against the unpruned models in Table \ref{tab:throughput}. CHEX method achieves $1.8\times\sim2.5\times$ actual runtime throughput accelerations on PyTorch with one NVIDIA V100 GPU in float32.}

\begingroup
\setlength{\tabcolsep}{4.7pt}
\begin{table}[h]
\centering
\footnotesize
\begin{tabular}{l | c c c | c c c}\toprule
    \textbf{Model} & \multicolumn{3}{c|}{ResNet-50} & \multicolumn{3}{c}{ResNet-101} \\\midrule
    \textbf{FLOPs reduction} & 0\% & 50\% & 75\% & 0\% & 50\% & 75\% \\\midrule
    \textbf{Throughput (img/s)} & 1328 & \textbf{2347} & \textbf{3259} & 840 & \textbf{1536} & \textbf{2032} \\\bottomrule
\end{tabular}
\caption{Comparison of the actual inference runtime throughput.}
\label{tab:throughput}
\end{table}
\endgroup

\subsection*{A2.9.~~~Comparison of training cost}
\begin{figure}[h]
\centering
\includegraphics[scale=0.23]{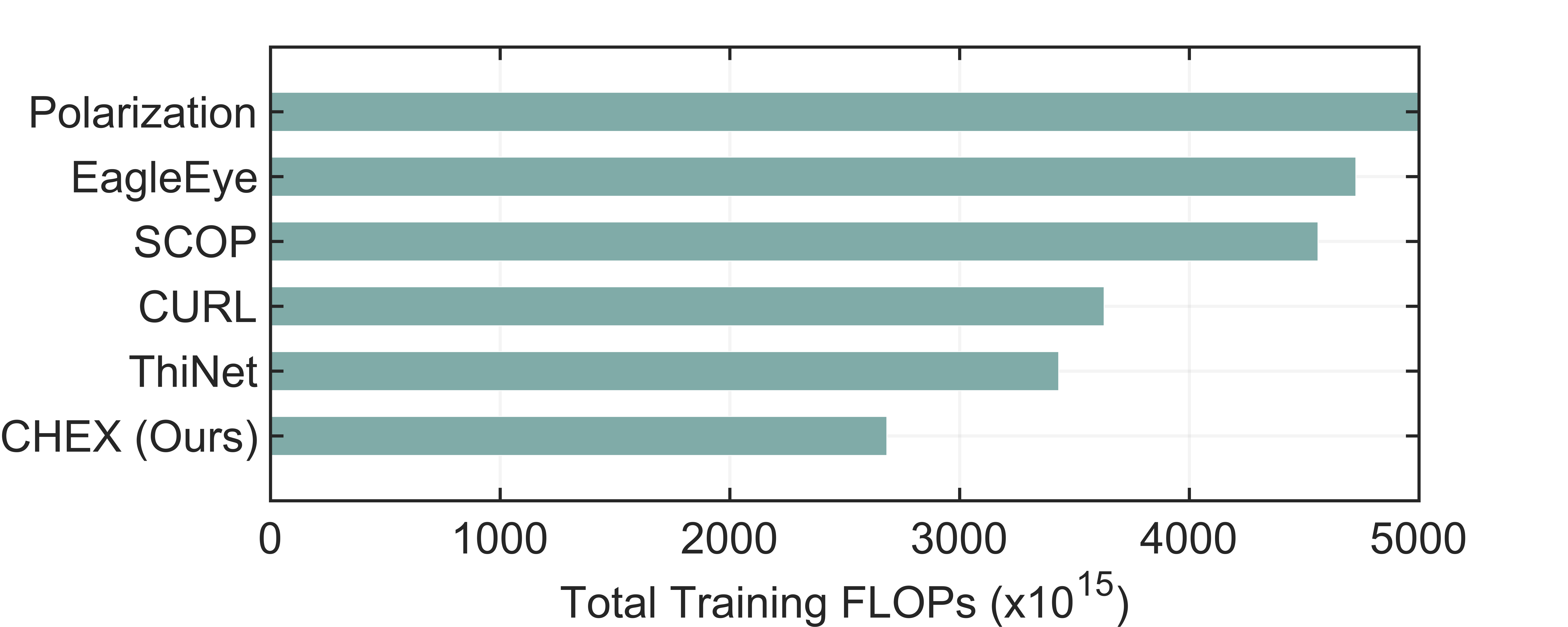}
\caption{Comparison of the total training FLOPs for pruning ResNet-50 on ImageNet.}
\label{fig:train_flops}
\end{figure}

{
In Figure \ref{fig:train_flops}, we compare the total training FLOPs of CHEX versus prior arts for pruning ResNet-50 on ImageNet. CHEX achieves higher accuracy at a fraction of the training cost. This is because CHEX obtains the sub-model in one training pass from scratch, circumventing the expensive pretrain-prune-finetune cycles.
}

\section*{A3.~~~Convergence analysis}

\subsection*{A3.1.~~~Problem formulation}
Training deep neural networks can be formulated into minimizing the following problem:
\begin{align}
\min_{\bW \in \mathbb{R}^d}\quad  F(\mathbf{W}) = \mathbb{E}_{x \sim \mathcal{D}}[f(\mathbf{W}; x)]
\end{align}
where $\mathbf{W} \in \mathbb{R}^d$ is the model parameter to be learned, and $f(\mathbf{W}; x)$ is a non-convex loss function in terms of $\mathbf{W}$. The random variable $x$ denotes the data samples that follow the distribution $\cD$. $\mathbb{E}_{x\sim \cD}[\cdot]$ denotes the expectation over the random variable $x$. 

\subsection*{A3.2.~~~Notation}
\label{app-notation}
We clarify several notions to facilitate the convergence analysis. 
\begin{itemize}
	\item $\bW_{t}$ denotes the complete model parameter at the $t$-th iteration.
	\item $m_t \in \RR^d$ is a mask vector at the $t$-th iteration.
	\item $x_{t}$ is the data sampled at the $t$-th iteration.
\end{itemize}

\subsection*{A3.3.~~~Algorithm formulation}
With notations introduced in the above subsection, the proposed CHEX method can be generalized in a way that is more friendly for convergence analysis, see Algorithm \ref{algorithm: math_friendly}. 

\begin{algorithm}[h]
	\small
	\textbf{Input}: Initialize $\bW_0$ and $m_0$ randomly \;
	\For{iteration $t=0,1,\cdots, T$}
	{
		Sample data $x_t$ from distribution $\mathcal{D}$ \;
		Generate a mask $m_t$ following some rules \;
		Update $\bW_{t+1} = \bW_{t} - \eta \nabla f(\bW_t \odot m_t; x_t) \odot m_t$
	}
	\textbf{Output}: The pruned model parameter $\bW_T \odot m_T$\;
	\caption{CHEX (A math-friendly version)}
	\label{algorithm: math_friendly}
\end{algorithm}

\begin{remark}
Note that Algorithm \ref{algorithm: math_friendly} is quite general. It does not specify the rule to generate the mask $m_t$. This implies the convergence analysis established in Sec.~\ref{app-convg} does not rely on what specific $m_t$ is utilized. In fact, it is even allowed for $m_t$ to remain unchanged during some period. 
\end{remark}

\subsection*{A3.4.~~~Assumptions}
We now introduce several assumptions on the loss function and the gradient noise that are standard in the literature.
\begin{assumption}[\sc Smoothness] 
	\label{ass-smooth}
	We assume $F(\bW)$ is $L$-smooth, i.e., it holds for any $\bW_1,\bW_2\in \mathbb{R}^d$ that 
	\begin{align}\label{global-L}
	\|\nabla F(\bW_1) - \nabla F(\bW_2)\| \le L \|\bW_1 - \bW_2\|
	\end{align}
	or equivalently,
	\begin{align}\label{global-L2}
	&\ F(\bW_1) - F(\bW_2) \nonumber \\
	\le &\ \langle \nabla F(\bW_2), \bW_1-\bW_2 \rangle + \frac{L}{2} \|\bW_1 - \bW_2\|^2
	\end{align}
\end{assumption}


\begin{assumption}[\sc Gradient noise] 
	\label{ass-gn}
	We assume 
	\begin{align}
	\mathbb{E}\{\nabla f(\bW; x_t)\} &= \nabla F(\bW),  \label{sn-1}\\
	\mathbb{E}\|\nabla f(\bW; x_t) - \nabla F(\bW)\|^2 &\le \sigma^2,\label{sn-2}
	\end{align}
	where $\sigma>0$ is a constent.
	Moreover, we assume the data sample $x_t$ is independent of each other for any $t$. 
\end{assumption}
This assumption implies that the stochastic filter-gradient is unbiased and has bounded variance.

\begin{assumption}[\sc Mask-incurred Error] 
	\label{ass-mask}
	It holds for any $\bW$ and $m_t$ that 
	\begin{align}\label{2396agba}
	\|\bW - \bW \odot  m_t \|^2 &\le \delta^2 \|\bW\|^2 \\
	\|\nabla F(\bW) - \nabla F(\bW) \odot  m_t \|^2 &\le \zeta^2 \|\nabla F(\bW)\|^2 \label{bnbz90871}
	\end{align}
	where constants $\delta \in [0,1]$ and $\zeta \in [0,1]$. 
\end{assumption}

With \eqref{bnbz90871}, we have 
\cross{\begin{align}
 &\ \zeta^2 \|\nabla F(\bW)\|^2 \nonumber \\
 \ge&\ \|\nabla F(\bW) - \nabla F(\bW) \odot  m_t \|^2 \nonumber \\
 \ge &\ (\|\nabla F(\bW)\| - \|\nabla F(\bW) \odot  m_t\|)^2
\end{align}
This together with $\|\nabla F(\bW) \odot  m_t\| \le \|\nabla F(\bW)\|$ implies 
\begin{align}\label{876baz}
	\|\nabla F(\bW) \odot  m_t\|^2 \ge (1-\zeta)^2 \|\nabla F(\bW)\|^2
\end{align}
for any $\bW$ and $m_t$.}
\yi{\begin{align}
 &\ \zeta \|\nabla F(\bW)\| \nonumber \\
 \ge&\ \|\nabla F(\bW) - \nabla F(\bW) \odot  m_t \| \nonumber \\
 \ge &\ \|\nabla F(\bW)\| - \|\nabla F(\bW) \odot  m_t\|
\end{align}
which implies 
\begin{align}\label{876baz}
	\|\nabla F(\bW) \odot  m_t\|^2 \ge (1-\zeta)^2 \|\nabla F(\bW)\|^2
\end{align}
for any $\bW$ and $m_t$.
}

\subsection*{A3.5.~~~Convergence analysis}
\label{app-convg}
Now we are ready to establish the convergence property. 
\begin{theorem}[\sc Convergence property]
Under Assumptions \ref{ass-smooth} -- \ref{ass-mask}, if learning rate $\eta = \frac{\sqrt{2 C_0}}{\sigma \sqrt{ L (T+1)}}$ in which $C_0 = \EE[F(\bW_0)]$, it holds that 
\begin{align}\label{znabnab2389776z}
&\ \frac{1}{T+1}\sum_{t=0}^T  \EE\|\nabla F(\bW_t\odot m_t)\|^2 \nonumber \\
\le&\ \frac{4\sigma\sqrt{L C_0}}{(1 \hspace{-0.5mm}-\hspace{-0.5mm} \zeta)^2\sqrt{T+1}} \hspace{-0.5mm} + \hspace{-0.5mm} \frac{2L^2\delta^2}{(T+1)(1-\zeta)^2} \sum_{t=0}^T \EE \| \bW_t\|^2
\end{align}
\end{theorem}

\begin{proof}
With inequality \eqref{global-L2} and Line 6 in Algorithm \ref{algorithm: math_friendly}, it holds that 
\begin{align}\label{global-L2-1}
&\ F(\bW_{t+1}) - F(\bW_t) \nonumber \\
\le &\ -\eta \langle \nabla F(\bW_t), [\nabla f(\bW_t\odot m_t;x_t)] \odot m_t \rangle \nonumber \\
&\ + \frac{\eta^2 L}{2} \|[\nabla f(\bW_t\odot m_t;x_t)] \odot m_t  \|^2
\end{align}
With Assumption \ref{ass-gn}, it holds that 
\begin{align}
&\ \mathbb{E} \langle \nabla F(\bW_t), \nabla f(\bW_t\odot m_t;x_t) \odot m_t \rangle \nonumber \\
\overset{\eqref{sn-1}}{=}&\ \mathbb{E}  \langle \nabla F(\bW_t), \nabla F(\bW_t\odot m_t) \odot m_t \rangle \nonumber \\
=&\ \mathbb{E} \langle  \nabla F(\bW_t) \odot m_t,  \nabla F(\bW_t \odot m_t ) \odot m_t  \rangle \nonumber \\
=& \frac{1}{2} \EE \|\nabla F(\bW_t)  \odot m_t\|^2 + \frac{1}{2} \EE \|\nabla F(\bW_t \odot m_t) \odot m_t\|^2 \nonumber \\
& \quad -  \frac{1}{2} \EE \|\nabla F(\bW_t)  \odot m_t - \nabla F(\bW_t \odot m_t)\odot m_t\|^2 \nonumber \\
\ge &\frac{1}{2} \EE \|\nabla F(\bW_t)  \odot m_t\|^2 + \frac{1}{2} \EE \|\nabla F(\bW_t \odot m_t) \odot m_t\|^2  \nonumber \\
& \quad - \frac{1}{2} \EE \|\nabla F(\bW_t)  - \nabla F(\bW_t \odot m_t) \|^2 \nonumber \\ 
\overset{\eqref{global-L}}{\ge} &\frac{1}{2} \EE \|\nabla F(\bW_t)  \odot m_t\|^2 + \frac{1}{2} \EE \|\nabla F(\bW_t \odot m_t) \odot m_t\|^2  \nonumber \\
& \quad - \frac{L^2}{2} \EE \| \bW_t  - \bW_t \odot m_t\|^2  \nonumber \\ 
\overset{\eqref{2396agba}}{\ge} &\frac{1}{2} \EE \|\nabla F(\bW_t)  \odot m_t\|^2 + \frac{1}{2} \EE \|\nabla F(\bW_t \odot m_t) \odot m_t\|^2  \nonumber \\
& \quad - \frac{L^2\delta^2}{2} \EE \| \bW_t \|^2 \nonumber \\
\overset{\eqref{876baz}}{\ge} &\frac{(1-\zeta)^2}{2} \EE \|\nabla F(\bW_t) \|^2 + \frac{(1-\zeta)^2}{2}  \EE \|\nabla F(\bW_t \odot m_t)\|^2  \nonumber \\
& \quad - \frac{L^2\delta^2}{2} \EE \| \bW_t \|^2 \label{zn248az650a}
\end{align}
Furthermore, 
with Assumption \ref{ass-gn}, it holds that 
\begin{align}\label{znznbabq09}
& \EE\|[\nabla f(\bW_t\odot m_t;x_t)] \odot m_t\|^2 \nonumber \\
\le\ & \EE\|\nabla f(\bW_t\odot m_t;x_t)\|^2 \nonumber \\
\overset{\eqref{sn-2}}{\le}&\ \EE \|\nabla F(\bW_t\odot m_t)\|^2  + \sigma^2
\end{align}
Substituting \eqref{zn248az650a} and \eqref{znznbabq09} into \eqref{global-L2-1}, we achieve 
\begin{align}\label{global-L2-2}
&\ \EE [F(\bW_{t+1}) - F(\bW_t)] \nonumber \\
\le &\ -\frac{\eta (1-\zeta)^2}{2}\EE\|\nabla F(\bW_t)\|^2 \nonumber \\
&\ -\frac{\eta (1-\zeta)^2}{2}\EE\|\nabla F(\bW_t \odot m_t)\|^2  \nonumber \\
&\ + \frac{\eta L^2 \delta^2}{2} \EE \| \bW_t\|^2 \nonumber \\
&\ + \frac{\eta^2 L}{2} \EE\|\nabla F(\bW_t\odot m_t)\|^2 + \frac{\eta^2 L \sigma^2}{2} \nonumber \\
\le &\ -\frac{\eta (1-\zeta)^2}{4}\EE\|\nabla F(\bW_t)\|^2 \nonumber \\
&\ -\frac{\eta (1-\zeta)^2}{4}\EE\|\nabla F(\bW_t \odot m_t)\|^2  \nonumber \\
&\ + \frac{\eta L^2 \delta^2}{2} \EE \| \bW_t\|^2 + \frac{\eta^2 L \sigma^2}{2}
\end{align}
where the last inequality holds by setting $\eta \le \frac{(1-\zeta)^2}{2L}$. The above inequality will lead to
\begin{align}
&\ \frac{1}{T+1}\sum_{t=0}^T \EE\|\nabla F(\bW_t)\|^2 + \EE\|\nabla F(\bW_t\odot m_t)\|^2 \nonumber \\
\le&\ \frac{4}{\eta (1-\zeta)^2 (T+1)}\EE[F(\bW_0)] \nonumber \\
&\ + \frac{2L^2\delta^2}{(1-\zeta)^2(T+1)} \sum_{t=0}^T \EE \| \bW_t\|^2 + \frac{2\eta L \sigma^2}{(1-\zeta)^2} \nonumber \\
\le&\ \frac{4\sigma\sqrt{L C_0}}{(1 \hspace{-0.5mm}-\hspace{-0.5mm} \zeta)^2\sqrt{T+1}} \hspace{-0.5mm} + \hspace{-0.5mm} \frac{2L^2\delta^2}{(T+1)(1-\zeta)^2} \sum_{t=0}^T \EE \| \bW_t\|^2
\end{align}
where $C_0 = \EE[F(\bW_0)]$ and the last equality holds when $\eta = \frac{\sqrt{2 C_0}}{\sigma \sqrt{ L (T+1)}}$. The above inequality will lead to \eqref{znabnab2389776z}. 
\end{proof}

\section*{A.4.~~~Societal impact}
Our method can effectively reduce the computation cost of diverse modern CNN models while maintaining satisfactory accuracy. This can facilitate the deployment of CNN models to real-world applications, such as pedestrian detection in autonomous driving and MRI image segmentation in clinic diagnosis. Moreover, our method does not increase the training cost compared to standard CNN model training. We provide a more affordable and efficient solution to CNN model compression, which is of high value for the community and society to achieve Green AI \cite{schwartz2020green}. On the other hand, our method cannot prevent the possible malicious usage, which may cause negative societal impact.

\section*{A.5.~~~Limitation}
CHEX tends to work better for more over-parameterized CNN models. When the model has substantial redundancy, our method can obtain efficient sub-models that recover the original accuracy well. \dl{On more compact or } \f{When compressing already}  under-parameterized CNN models, our method will \f{still} have \dl{meaningful} \f{noticeable} accuracy loss\f{, though such loss may still be less than the comparable methods (See Table~\ref{tab:lightweight_cnns} in Appendix~A2.1}).

CHEX is primarily evaluated on diverse computer vision (CV) tasks in this paper.
More evaluations are required to verify the broader applicability of CHEX to other domains, such as natural language processing, and we leave it as one of our future works.

Finally, CHEX \dl{uses a relatively ad-hoc technique to} reflect\f{s} the layer importance based on the scaling factors in \f{the} batch-norm layers. Although this technique can provide meaningful guidance in allocating the number of channels in the sub-models, deeper understanding on why this mechanism works is still an open question.



\end{document}